\documentclass[10pt,twocolumn,letterpaper]{article}

\usepackage{wacv}              

\usepackage{graphicx}
\usepackage{booktabs}
\usepackage{comment}
\usepackage{pgfplots}
\pgfplotsset{compat=1.18}
\usepackage{pgfplotstable}
\usepgfplotslibrary{groupplots}
\usepackage{pgfplotstable}
\usepgfplotslibrary{statistics} 
\usepackage{multirow}
\usepackage{tikz}
\newcommand\ours{\texttt{TTA-CaP}\xspace}
\usepackage{pifont}

\usepackage[table]{xcolor}
\usepackage{pifont}
\usepackage{wrapfig}
\usepackage{amsmath,amssymb} 
\usepackage[ruled,vlined]{algorithm2e}
\usepackage{siunitx}
\nocite{*}

\definecolor{pinkNonTTA}{RGB}{255,248,251} 
\definecolor{pinkTTA}{RGB}{255,240,246}    

\definecolor{coral}{RGB}{255,127,80}
\definecolor{teal}{RGB}{0,128,128}
\definecolor{violet}{RGB}{138,43,226}
\definecolor{gold}{RGB}{255,191,0}

\definecolor{biovidcolor}{RGB}{213,94,0}
\definecolor{stresscolor}{RGB}{0,114,178}
\definecolor{bahcolor}{RGB}{0,158,115}

\newcommand\biovid{\texttt{BioVid}\xspace}
\newcommand\stressid{\texttt{StressID}\xspace}
\newcommand\bah{\texttt{BAH}\xspace}

\newcommand\war{\texttt{WAR}\xspace}
\newcommand\fonescore{\texttt{F1}\xspace}

\newcommand\gflops{\texttt{GFLOPs}\xspace}

\definecolor{wacvblue}{rgb}{0.21,0.49,0.74}
\usepackage[pagebackref,breaklinks,colorlinks,allcolors=wacvblue]{hyperref}

\def\wacvPaperID{1204} 
\def\confName{WACV}
\def\confYear{2027}

\title{Test-Time Adaptation via Cache Personalization for Facial Expression Recognition in Videos}

\author{ Masoumeh Sharafi\textsuperscript{1}, Muhammad Osama Zeeshan\textsuperscript{1}, Soufiane Belharbi\textsuperscript{1},\\ Alessandro Lameiras Koerich\textsuperscript{2}, Marco Pedersoli\textsuperscript{1}, Eric Granger\textsuperscript{1}\\[4pt] \textsuperscript{1}LIVIA, Department of Systems Engineering, \'{E}TS Montreal, Canada\\ \textsuperscript{2}LIVIA, Department of Software and IT Engineering, \'{E}TS Montreal, Canada\\[3pt] {\tt\small \{masoumeh.sharafi.1,muhammad-osama.zeeshan.1\}@ens.etsmtl.ca}\\ {\tt\small \{soufiane.belharbi,marco.pedersoli,alessandro.koerich,eric.granger\}@etsmtl.ca} }

\begin{document}
\maketitle
\begin{abstract}
Facial expression recognition (FER) in videos requires model personalization to capture the considerable variations across subjects. Vision-language models (VLMs) offer strong transfer to downstream tasks through image-text alignment, but their performance can still degrade under inter-subject distribution shifts. Personalizing models using test-time adaptation (TTA) methods can mitigate this challenge. However, most state-of-the-art TTA methods rely on unsupervised parameter optimization, introducing computational overhead that is impractical in many real-world applications. 
This paper introduces TTA through Cache Personalization (\ours), a cache-based TTA method that enables cost-effective (gradient-free) personalization of VLMs for video FER. Prior cache-based TTA methods rely solely on dynamic memories that store test samples, which can accumulate errors and drift due to noisy pseudo-labels. 
To address this limitation, \ours introduces three complementary caches -- a personalized static cache constructed through feature-statistics matching, a positive target cache that accumulates reliable subject-specific samples, and a negative target cache that stores low-confidence cases as negative samples. 
To prevent target-cache corruption, a tri-gate mechanism controls cache updates based on temporal stability, confidence, and consistency with the personalized static cache.
Together, the caches provide complementary, subject-matched positive and negative evidence for robust personalization.
Finally, \ours refines predictions by fusing embeddings, yielding representations that support temporally stable video-level predictions. 
Our experiments on three challenging video FER datasets — \biovid, \stressid, and \bah — indicate that \ours can outperform state-of-the-art TTA methods under subject-specific and environmental shifts, while maintaining low computational and memory overhead for real-world deployment. Code is available at \href{https://github.com/MasoumehSharafi/TTA-CaP}{https://github.com/MasoumehSharafi/TTA-CaP}.
\end{abstract}
    
\section{Introduction} \label{sec:intro}

Video-based FER is a key component of affective computing applications, such as human--computer interaction~\cite{pu2023convolutional}, health monitoring~\cite{gayamorey2025deeplearningbasedfacialexpression}, and clinical assessment of pain, depression, and stress~\cite{calvo2010affect}. Unlike image-based FER, video FER must account for both facial appearance and the temporal evolution of expressions. Expressions may develop gradually, remain subtle across several frames, or involve brief transitions between neutral and expressive states. Consequently, individual frames are not equally informative, and predictions may fluctuate within the same video. These challenges are amplified in subject-independent settings by variations in facial morphology, expression intensity, behavioral patterns, and acquisition conditions. Models trained on a fixed group of subjects may therefore generalize poorly to unseen individuals, motivating models that can transfer effectively while adapting to subject-specific expression patterns.


\begin{figure*}[t!]
  \centering
  \includegraphics[width=0.99\linewidth, height=0.21\linewidth]{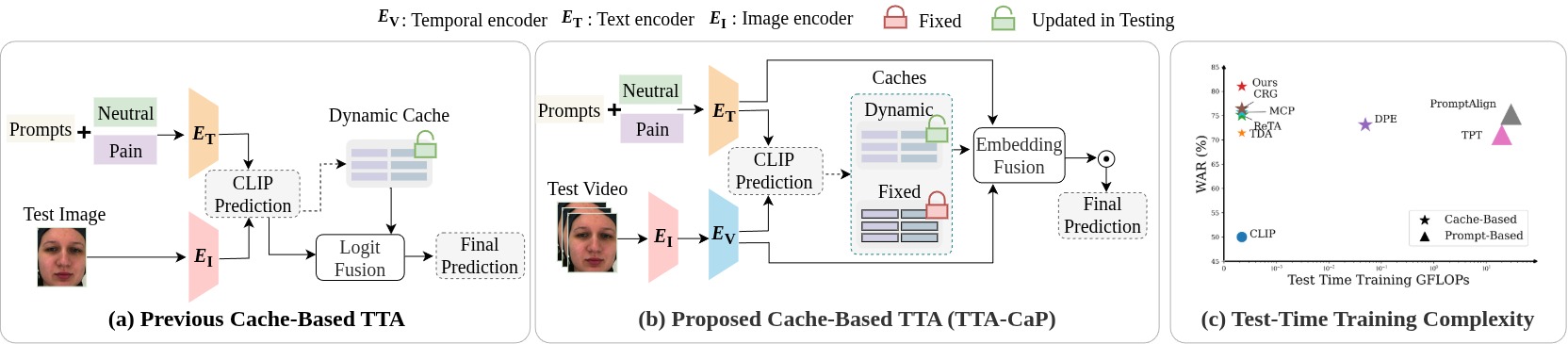}
  \caption{A comparison of SOTA cache-based TTA methods (\textbf{a}), which store pseudo-labeled test embeddings only in dynamic target caches and perform logit-level fusion, with our \ours (\textbf{b}). \ours adapts a VLM to a subject video at test time by combining a personalized static cache with dynamic target-domain caches, and refines predictions through embedding-level fusion. \textbf{(c)} Average weighted average recall (\war) versus \gflops for test-time adaptation, over target subjects of the \biovid dataset. Marker shapes distinguish cache- vs.\ prompt-based methods, while marker sizes indicate runtime (ms) at test time.
  }
  \label{fig:Motivation}
\end{figure*}

Large-scale VLMs such as CLIP~\cite{radford2021learning} have shown strong transfer to downstream tasks through aligned image--text representations learned from web-scale data~\cite{yu2022coca, li2021supervision, li2023blip}. However, their performance in video FER may decline under test-time domain shifts caused by subject and acquisition variations~\cite{shu2023clipood,tu2023closer}. This motivates model \emph{personalization}, i.e., adaptation to the subtle expression patterns of an unseen subject~\cite{zeeshan2024subject, sharafi2025disentangled, sharafi2025personalized, zeeshan2025musaco, zeeshan2025progressive}. Although CLIP has been adapted to FER, existing methods such as EmoCLIP~\cite{foteinopoulou2024emoclip}, DFER-CLIP~\cite{zhao2023prompting}, and Exp-CLIP~\cite{zhao2025enhancing} mainly improve representations, prompting, or temporal modeling, rather than subject-level test-time adaptation.

Personalization of VLMs through test-time adaptation (TTA) remains underexplored for video FER. Prompt-based TTA methods update prompts or lightweight modules at test time using unsupervised objectives such as entropy minimization and augmentation consistency~\cite{shu2022test, feng2023diverse, abdul2023align, yoon2024c}. While effective, they require repeated backpropagation at test time, increasing computational cost.

Cache-based TTA provides a gradient-free alternative by storing and retrieving test embeddings to refine predictions~\cite{karmanov2024efficient, liang2025advancing, zhang2024dual, zhu2025dynamic, chen2025multi, zhai2025mitigating}. Representative methods such as TDA~\cite{karmanov2024efficient}, MCP~\cite{chen2025multi}, and CRG~\cite{zhai2025mitigating} improve adaptation through dynamic caches or prototype-based representations, but do not construct a source-derived cache personalized to each target subject. In contrast, \ours uses target feature statistics to select subject-relevant source prototypes as a personalized static prior, and complements this prior with positive and negative target caches. This distinction is particularly important for video FER, where ambiguous or transitional frames can produce temporally correlated pseudo-label errors that propagate through dynamic cache updates. To limit such errors, \ours further controls cache updates using temporal stability, prediction confidence, and consistency with the personalized prior, thereby distinguishing reliable subject-specific observations from uncertain samples.

In this paper, we propose a cache-based gradient-free TTA method, called TTA via Cache Personalization (\ours), for cost-effective personalization of VLMs for video-based FER. \ours performs test-time adaptation independently for each target-subject video. As illustrated in Fig.~\ref{fig:Motivation}(a), prior CLIP cache-based TTA methods target independent images and rely mainly on dynamic caches updated with pseudo-labels. In videos, ambiguous or transitional frames can introduce noisy pseudo-labels that propagate through subsequent cache updates. In contrast, as shown in Fig.~\ref{fig:Motivation}(b), \ours combines three caches: a personalized static cache, a positive cache of reliable target samples, and a negative cache of uncertain samples used as negative evidence. The static cache matches target-subject feature statistics with precomputed source-subject statistics to retrieve relevant source embeddings as a personalized prior. To update the two dynamic target caches, \ours introduces a tri-gate mechanism based on (i) temporal stability across recent frames, (ii) prediction confidence by entropy, and (iii) consistency with the personalized static cache. The retrieved representations are combined with the CLIP visual embedding through embedding-level fusion before classification. Unlike logit-level fusion, embedding-level fusion integrates retrieved evidence directly into the CLIP representation before classification, supporting temporal aggregation. Fig.~\ref{fig:Motivation}(c) further shows that \ours improves the accuracy--efficiency trade-off without the test-time optimization required by prompt-tuning methods.

Our contributions are summarized as follows. 
\textbf{(1)} We propose \ours, a cache-based, gradient-free TTA method, which couples dynamic target caches with a personalized static cache, constructed by matching the feature statistics of the target subject to precomputed subject-level statistics from source and selecting the most relevant expression representations, to mitigate noisy pseudo-label updates and refine predictions for personalized video FER. 
\textbf{(2)} A tri-gate mechanism is introduced that controls target-cache updates using temporal stability, prediction confidence, and consistency with the personalized cache, together with embedding-level fusion that combines CLIP visual embeddings with cache-retrieved representations for reliable video-level prediction.
\textbf{(3)} An extensive set of experiments conducted on three challenging video FER benchmarks, \biovid~\cite{walter2013biovid} (pain estimation), \stressid~\cite{chaptoukaev2023stressid} (stress recognition), and \bah~\cite{gonzalez2025bah} (ambivalence-hesitancy recognition). Results show that \ours outperforms state-of-the-art prompt-tuning and cache-based methods.

\section{Related Work}
\label{sec:Related_work}

\noindent\textbf{Vision-Language Models for FER.}
VLMs trained with large-scale image--text contrastive learning, such as CLIP, enable prompt-based recognition by reframing classification as image--text matching~\cite{radford2021learning}. This is appealing for FER because expression classes are semantic; however, FER is fine-grained and entangled with identity and capture conditions, making performance sensitive to prompt design and inter-subject shifts in videos. Recent CLIP-based FER methods improve language guidance and temporal modeling. CLIPER uses multiple expression descriptors to learn more discriminative static and dynamic representations~\cite{li2024cliper}, while DFER-CLIP incorporates temporal modeling and enriched textual descriptions to better align subtle dynamics with language priors~\cite{zhao2023prompting}. In zero-shot settings, EmoCLIP uses natural-language descriptions to improve generalization~\cite{foteinopoulou2024emoclip}, and Exp-CLIP transfers task knowledge distilled from LLMs~\cite{zhao2025enhancing}. Multimodal prompt alignment and parameter-efficient adaptation further improve dynamic FER~\cite{ma2025mpafer}. However, these methods mainly focus on prompting and training-time adaptation, leaving reliable test-time adaptation for videos underexplored.

\noindent\textbf{Test-Time Adaptation.}
Test-time adaptation (TTA) mitigates distribution shift by adapting a pretrained model using only unlabeled test data during inference. Early methods optimize auxiliary self-supervised or entropy-based objectives, as in test-time training, Tent, and MEMO~\cite{sun2020test,wang2020tent,zhang2022memo}. More recent approaches, including EATA, SAR, CoTTA, and RoTTA, improve stability under noisy, small-batch, or non-stationary test streams through selective updates, robust optimization, and memory mechanisms~\cite{niu2022efficient,niu2023sar,wang2022continual,yuan2023robust}. However, these methods still rely on iterative parameter updates and may suffer from confirmation bias or limited compatibility with LayerNorm-based backbones.

TTA for VLMs instead exploits CLIP's aligned image--text space and strong zero-shot priors. Methods such as TPT and PromptAlign adapt textual prompts using entropy minimization, augmented views, or feature-statistics alignment~\cite{shu2022test,abdul2023align}. Other approaches employ prompt ensembles, weight averaging, or CLIP-specific objectives to reduce pseudo-label drift and preserve image--text alignment~\cite{osowiechi2024watt,lafon2025cliptta}. Despite their effectiveness, these methods require repeated backpropagation at test time and remain sensitive to optimization choices and unreliable predictions, motivating gradient-free alternatives.

\noindent\textbf{Cache-Based TTA Methods.}
Cache-based TTA methods adapt predictions using stored target features or prototypes in the embedding space~\cite{karmanov2024efficient}. TDA~\cite{karmanov2024efficient}, DPE~\cite{zhang2024dual}, and ReTA~\cite{liang2025advancing} improve cache reliability through dynamic updates and prototype refinement, while MCP~\cite{chen2025multi} and CRG~\cite{zhai2025mitigating} exploit multiple prototypes or positive/negative evidence. In contrast, TTA-CaP is designed for subject-based video FER and combines a personalized static source cache, dynamic positive/negative target caches, temporal consistency, tri-gated updates, and embedding-level fusion. Despite these advances, existing cache-based TTA is not tailored to subject-based video FER: pseudo-label-driven target memories can degrade under frame-level noise, and most methods do not exploit subject-specific source priors to anchor class semantics, while long videos also raise memory and retrieval-cost concerns.
\section{TTA through Cache Personalization}
\label{sec:methodology}

\begin{figure*}[t!]
  \centering
  \includegraphics[width=0.9\linewidth]{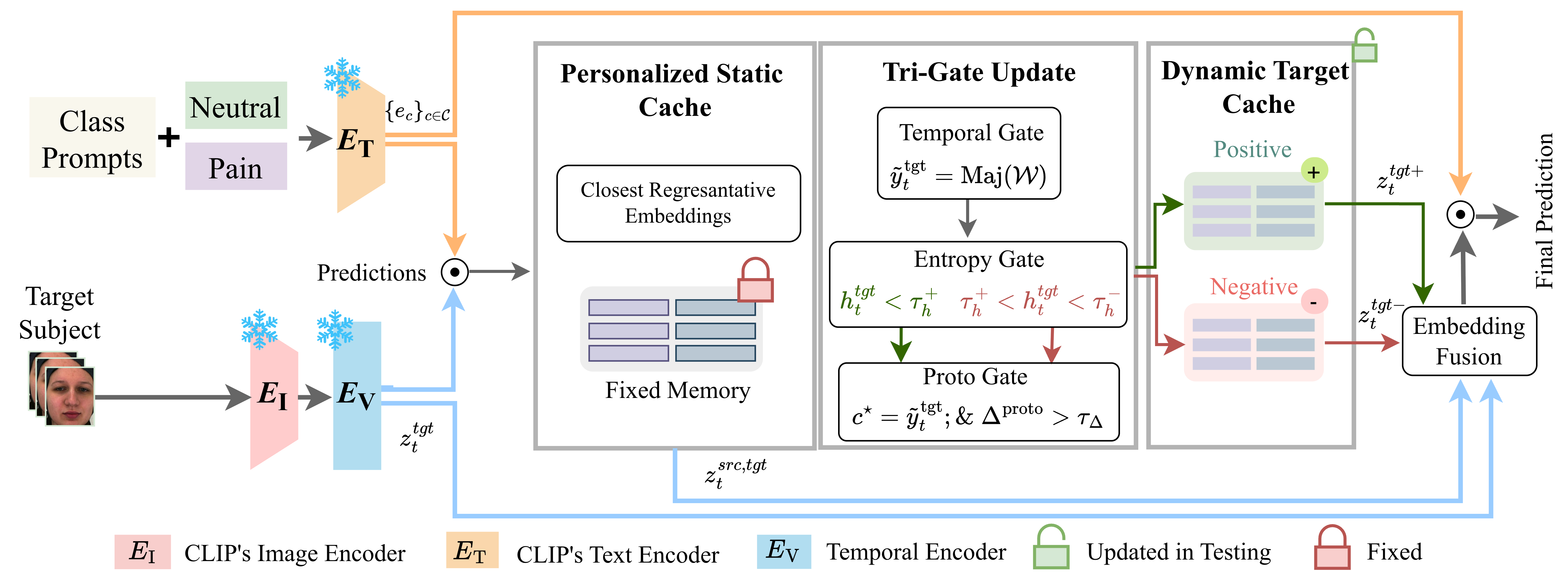}
\caption{Overview of \ours. At test time, frozen CLIP encoders produce visual and text embeddings and an initial prediction through image--text similarity. A personalized static cache provides stable subject-relevant references. While, a tri-gate evaluates temporal stability, prediction confidence, and consistency with the static cache to control updates to the positive and negative target caches. This design limits the accumulation and propagation of noisy pseudo-labels. Final predictions are obtained through embedding-level fusion of the temporally aggregated representation with retrieved cache representations, enabling cost-effective personalization without test-time training.}

  \label{fig:Main_fig}
\end{figure*}

The proposed \ours method reduces noisy pseudo-labels at test time through a personalized static cache as a subject-specific reference, a positive target cache for reliable samples, and a negative target cache for uncertain evidence. At each time step, the current representation is used for initial prediction, retrieval from previously stored cache entries, and embedding-level fusion. The refined embedding is then compared with the class text embeddings using cosine similarity, after which the tri-gate determines whether the current sample is added to a dynamic cache, preventing self-retrieval.

Fig.~\ref{fig:Main_fig} summarizes the framework. Let $\mathrm{tgt}$ denote a target subject and $\mathcal{X}^{\mathrm{tgt}}=\{\bold{X}^{\mathrm{tgt}}_t\}_{t=1}^{T}$ be an input video sequence of length $T$, where
$\bold{X}^{\mathrm{tgt}}_t \in \mathbb{R}^{H \times W \times 3}$ denotes the RGB frame at time $t$.
An image encoder $E_{\mathrm{I}} :\mathbb{R}^{H \times W \times 3} \rightarrow\mathbb{R}^{d}$ maps each frame to a $d$-dimensional embedding: $\bold{v}^{\mathrm{tgt}}_t=E_{\mathrm{I}}(\bold{X}^{\mathrm{tgt}}_t)$.
To model short-range temporal dynamics, we define a temporal encoder $E_{\mathrm{V}} : (\mathbb{R}^{d})^{L}\rightarrow\mathbb{R}^{d}$, which operates on a sliding window of length $L$. For $t \ge L$, the temporally aggregated representation is
$\bold{z}^{\mathrm{tgt}}_t =E_{\mathrm{V}}\big(\bold{v}^{\mathrm{tgt}}_{t-L+1},\dots,\bold{v}^{\mathrm{tgt}}_{t}\big)$. The representation $\bold{z}^{\mathrm{tgt}}_t$ is $\ell_2$-normalized before prediction and cache retrieval.
Let $\mathcal{C}$ denote the finite set of class labels. For each $c\in\mathcal{C}$, let $p_c$ be a predefined textual prompt. CLIP text encoder $E_{\mathrm{T}} :\mathcal{P}\rightarrow\mathbb{R}^{d}$ produces normalized class embeddings
$\bold{e}_c$.
Given frozen CLIP encoders, classification logits at time $t$ can be obtained via cosine similarity $\ell^{\mathrm{tgt}}_{t,c}=\bold{z}^{\mathrm{tgt}\top}_t\bold{e}_c$. The predicted label is $\tilde{y}^{\mathrm{tgt}}_t=\displaystyle\arg\max_{c \in \mathcal{C}}\ell^{\mathrm{tgt}}_t(c)$.

\subsection{Personalized Static Cache}
\label{sec:src_cache}

A personalized static cache provides a stable reference for retrieval and reliability assessment during test-time adaptation. Before deployment, the frozen image encoder $E_{\mathrm{I}}$ extracts class-conditional source embeddings and subject-level feature statistics. To construct a compact reference pool, density-based spatial clustering of applications with noise (DBSCAN)~\cite{ester1996density} is applied independently to each subject--class subset. DBSCAN identifies dense feature regions while excluding isolated, potentially unrepresentative samples. Its neighborhood radius and minimum-neighbor parameters are selected per subset to account for inter-subject variability and class imbalance. Candidate configurations are evaluated using the outlier ratio, avoidance of degenerate solutions, and stability under bootstrap resampling, measured by the mean adjusted Rand index (ARI)~\cite{rand1971objective,hubert1985comparing,vinh2010information,ben2001stability}. For each cluster, we select the observed embedding closest to its centroid as the representative, forming a compact class-wise pool of representative embeddings. The associated subject-level statistics are stored to guide target-specific personalization. At test time, the reference pool is personalized once for each target subject through statistics-guided selection. The mean and diagonal covariance of all unlabeled target embeddings are compared with the precomputed statistics of each source subject using the diagonal Fr\'echet distance. Source subjects are ranked by increasing distance, and the reference embeddings from the top-$m$ matched subjects are retrieved. Their class-wise embeddings form the personalized static cache $\mathcal{P}^{\mathrm{tgt}}_c \subset \mathbb{R}^{d},\; c \in \mathcal{C}$. The cache remains fixed during test-time adaptation and supports retrieval and reliability assessment. No raw source samples are accessed at test time; only stored statistics and reference embeddings are required.

\subsection{Dynamic Target Caches}
\label{sec:tar_cache}
At test time, two initially empty caches are maintained for each target-subject video: a positive cache for reliable samples and a negative cache for ambiguous samples. After constructing the personalized static cache using the unlabeled target video, temporal representations are processed sequentially to update the dynamic target caches. During this stage, each cache update uses only the current and previously processed temporal representations. 

\noindent\textbf{Positive cache.}
For each target subject, a limited-capacity positive cache $C^{\text{tgt}+}(t)=\{(\bold{z}^\text{tgt}_k, \tilde{y}^\text{tgt}_k, h^\text{tgt}_k)\}_{k \in \mathcal{I}^{+}_t}$ stores reliable target samples at time $t$, where $\bold{z}_k \in \mathbb{R}^{d}$ denotes a reliable pseudo-labeled embedding, $\tilde{y}_k \in \mathcal{C}$ is its pseudo-label, $h^\text{tgt}_k$ is the predictive uncertainty (entropy) measured at the time of insertion, and $\mathcal{I}^{+}_t \subseteq \{1,\dots,t\}$ indexes the insertion times. This cache is maintained per class with a fixed capacity to keep memory constant. When the capacity of a class-specific cache is exceeded, the least reliable entries (highest entropy) are removed, so the cache preserves high-confidence examples over time.

\noindent\textbf{Negative cache.}
For each target subject, a limited-capacity negative cache
$C^{\mathrm{tgt}-}(t)=
\{(\boldsymbol{z}^{\mathrm{tgt}}_k,
\tilde{y}^{\mathrm{tgt}-}_k,
h^{\mathrm{tgt}}_k)\}_{k\in\mathcal{I}^{-}_t}$
stores moderately uncertain target samples as negative evidence.
Unlike the positive cache, its class index does not represent the
predicted class. Instead,
$\tilde{y}^{\mathrm{tgt}-}_k$ is the least-likely class of sample $k$;
thus, $C^{\mathrm{tgt}-}_c$ contains uncertain samples for which class
$c$ received the lowest prediction score.

For a current sample with pseudo-label $c$, we query
$C^{\mathrm{tgt}-}_c$. Hence, the retrieved negative samples are not
required to have pseudo-label $c$; rather, they provide negative
evidence for $c$. Because these samples are moderately uncertain,
they may contain evidence for several classes, but the evidence for
$c$ is, by construction, the weakest. Their subtraction during fusion
is therefore intended to reduce non-$c$ evidence relatively more than
evidence for $c$, increasing the class margin in favor of the current
prediction. Each class has a fixed capacity; when exceeded, the
highest-entropy entry is removed.


\subsection{Test-Time Adaptation}
\label{sec:TTA}
\ours operates independently on each target-subject video, performing sequential adaptation throughout the video. At each time step $t$, the temporal representation $z_t^{\mathrm{tgt}}$ is obtained and compared with the class text embeddings $\{e_c\}_{c\in\mathcal{C}}$ to produce the initial CLIP prediction. The caches are then queried using $z_t^{\mathrm{tgt}}$, and the retrieved representations are fused with it to obtain the final prediction. Cache updates are then performed to store target samples with their pseudo-labels. Cache updates are controlled by a tri-gate mechanism, ensuring that only reliable embeddings are admitted to the positive or negative target cache. \textbf{Temporal gating:}
Predictions in video FER can fluctuate due to motion, blur, or subtle expressions. To prevent cache updates from being triggered by transient fluctuations, the temporal gate enforces short-term consistency: an update is accepted only if the current predicted label matches the majority label over the last $\mathcal{W}$ frames, i.e., $\tilde{y}^{\mathrm{tgt}}_t=\mathrm{Maj}(\tilde{y}^{\mathrm{tgt}}_{t-\mathcal{W}+1:t})$. \textbf{Entropy gating:}
Prediction confidence is measured using the entropy of the predicted class distribution, where lower entropy indicates higher confidence. Two thresholds, $\tau_h^{+}$ and $\tau_h^{-}$, separate confident from moderately uncertain predictions. Samples with $h^{\mathrm{tgt}}_t<\tau_h^{+}$ are considered reliable and are eligible for the positive cache, while samples with $\tau_h^{+}\leq h^{\mathrm{tgt}}_t<\tau_h^{-}$ are considered moderately uncertain and are eligible for the negative cache, where they are indexed by their least-likely class. Predictions with $h^{\mathrm{tgt}}_t\geq\tau_h^{-}$ are considered too uncertain and are excluded from cache updates. This filtering prevents highly ambiguous samples from entering the caches while retaining moderately uncertain samples as class-specific negative evidence. \textbf{Prototype gating:}
The personalized static cache provides class prototypes for the target subject. For the current representation, class-wise similarity scores are computed using cosine similarity averaged over the top-$k$ nearest prototypes. Let $c^{\star}$ denote the class with the highest prototype similarity and $\Delta_{\mathrm{proto}}$ the margin between the highest and second-highest scores. The gate passes only if $c^{\star}=\tilde{y}^{\mathrm{tgt}}_t$ and $\Delta_{\mathrm{proto}}>\tau_{\Delta}$.

\noindent\textbf{Cache retrieval.}
Before updating the dynamic caches, the current representation
$\boldsymbol{z}^{\mathrm{tgt}}_t$ with pseudo-label
$c=\tilde{y}^{\mathrm{tgt}}_t$ retrieves evidence from three
class-specific cache partitions. From the personalized static cache
$\mathcal{P}^{\mathrm{tgt}}_c$, it retrieves source prototypes of class
$c$, while from the positive cache $\mathcal{C}^{\mathrm{tgt}+}_c$, it
retrieves reliable target samples whose pseudo-label is $c$.
In contrast, $\mathcal{C}^{\mathrm{tgt}-}_c$ contains moderately
uncertain samples for which $c$ was the \emph{least-likely} class.
Thus, the same index $c$ denotes positive evidence for $c$ in the
static and positive caches, but negative evidence for $c$ in the
negative cache. Within each selected partition, the $k$ embeddings
most cosine-similar to $\boldsymbol{z}^{\mathrm{tgt}}_t$ are retrieved.
For a nonempty partition $\mathcal{M}_c$, let
$k_{\mathcal{M}}=\min(k,|\mathcal{M}_c|)$ and
$R(\boldsymbol{z},\mathcal{M}_c)=
\frac{1}{k_{\mathcal{M}}}\sum_{i\in
\mathcal{N}_{k_{\mathcal{M}}}(\boldsymbol{z},\mathcal{M}_c)}
\boldsymbol{m}_i$, where $\mathcal{N}_{k_{\mathcal{M}}}$ contains the
most cosine-similar entries; if $\mathcal{M}_c=\varnothing$,
$R(\boldsymbol{z},\mathcal{M}_c)=\mathbf{0}$.
Accordingly,
$\boldsymbol{z}^{\mathrm{src,tgt}}_t=
R(\boldsymbol{z}^{\mathrm{tgt}}_t,\mathcal{P}^{\mathrm{tgt}}_c)$,
$\boldsymbol{z}^{\mathrm{tgt}+}_t=
R(\boldsymbol{z}^{\mathrm{tgt}}_t,\mathcal{C}^{\mathrm{tgt}+}_c)$,
and $\boldsymbol{z}^{\mathrm{tgt}-}_t=
R(\boldsymbol{z}^{\mathrm{tgt}}_t,\mathcal{C}^{\mathrm{tgt}-}_c)$.


\noindent\textbf{Cache update.}
\label{sec:cache_update}
Since pseudo-labels can be noisy, cache updates are restricted to reliable cases: the temporal check avoids storing momentary prediction flips, and the prototype check avoids storing samples that disagree with the fixed source cache. When both checks pass, the entropy thresholds determine how the sample is stored. If the entropy is below $\tau_h^{+}$, the current embedding $\bold{z}^{\mathrm{tgt}}_t$ is added to the positive cache with its predicted label, treating it as reliable target information. If the entropy lies between $\tau_h^{+}$ and $\tau_h^{-}$, $\bold{z}^{\mathrm{tgt}}_t$ is added to the negative cache with a negative pseudo-label from that uncertain prediction, so ambiguous cases do not enter the positive cache. If the entropy is above $\tau_h^{-}$, no cache update is performed.

\noindent\textbf{Embedding-level fusion.}
\label{sec:fusion}
Using the retrieved representations, fusion is performed directly in the shared CLIP embedding space rather than at the logit level, allowing cache evidence to refine the visual representation before image-text classification. For each frame $t$, the current temporal representation is combined with representations retrieved from the personalized static and dynamic target caches:
\begin{equation}
\boldsymbol{z}^{\text{tgt,fuse}}_t
=
\boldsymbol{z}^{\text{tgt}}_t
+\boldsymbol{z}^{\text{src,tgt}}_t
+\boldsymbol{z}^{\text{tgt}+}_t
-\boldsymbol{z}^{\text{tgt}-}_t,
\end{equation}
where $\boldsymbol{z}^{\text{src,tgt}}_t$ is retrieved from the fixed personalized source cache, $\boldsymbol{z}^{\text{tgt}+}_t$ from the positive target cache, and $\boldsymbol{z}^{\text{tgt}-}_t$ from the negative target cache.

For each class $c\in\mathcal{C}$, we precompute a text embedding using the CLIP text encoder $E_{\text{T}}$ and a class prompt $\tau_c$:
\begin{equation}
\boldsymbol{e}_c = \frac{E_{\text{T}}(\tau_c)}{\|E_{\text{T}}(\tau_c)\|_2}.
\end{equation}
Given the fused frame embedding $\boldsymbol{z}^{\text{tgt,fuse}}_t$, frame-level logits are obtained by cosine similarity to the text embeddings:
\begin{equation}
\ell_t(c)=\eta\,\cos\!\Big(\frac{\boldsymbol{z}^{\text{tgt,fuse}}_t}{\|\boldsymbol{z}^{\text{tgt,fuse}}_t\|_2},\,\boldsymbol{e}_c\Big),
\qquad
\hat{y}_t=\arg\max_{c\in\mathcal{C}}\ell_t(c),
\end{equation}
where $\eta$ is the CLIP logit scale.

For a video of $T$ frames, we aggregate frame-level logits by temporal averaging and predict the final label by argmax:
\begin{equation}
\bar{\ell}(c)=\frac{1}{T}\sum_{t=1}^{T}\ell_t(c),
\qquad
\hat{y}=\arg\max_{c\in\mathcal{C}}\bar{\ell}(c).
\end{equation}

\vspace{-0.4cm}

\begin{table*}[t!]
\centering

\scriptsize
\resizebox{0.65\linewidth}{!}{%
\begin{tabular}{l|l|cc|cc|cc}
\toprule
\multicolumn{1}{l|}{\multirow{2}{*}{\textbf{Protocol}}} &
\multicolumn{1}{l|}{\multirow{2}{*}{\textbf{Method}}} &
\multicolumn{2}{c|}{\textbf{BioVid}} &
\multicolumn{2}{c|}{\textbf{StressID}} &
\multicolumn{2}{c}{\textbf{BAH}} \\
\cmidrule(lr){3-4} \cmidrule(lr){5-6} \cmidrule(lr){7-8}
& & \war $\uparrow$ & \fonescore $\uparrow$ & \war $\uparrow$ & \fonescore $\uparrow$ & \war $\uparrow$ & \fonescore $\uparrow$ \\
\midrule

ZS & CLIP-ViT-B/32{\fontsize{4}{12} \selectfont (ICML'21)} & 50.0 & 33.3 & 60.4 & 34.8 & 39.5 & 28.1 \\
\midrule

 
& CLIP-ViT-B/32\textsuperscript{$\dagger$}{\fontsize{4}{12} \selectfont (ICML'21)} & 69.7 & 66.6 & 67.0 & 44.5 & 60.4 & 39.8 \\

& EmoCLIP~\cite{foteinopoulou2024emoclip}{\fontsize{4}{12} \selectfont (FG'24)} & 67.7 & 63.4 & 63.5 & 35.9 & 56.2 & 36.5 \\

& X-CLIP~\cite{ni2022expanding}{\fontsize{4}{12} \selectfont (ECCV'22)} & 70.9 & 57.9 & 62.3 & 41.3 & 63.0 & 39.2 \\

\multirow{-4}{*}{FT} & Exp-CLIP~\cite{zhao2025enhancing} {\fontsize{4}{12} \selectfont (WACV'25)}& 70.2 & 66.7 & 63.1 & 44.5 & 62.2 & 38.5 \\
\midrule


& TPT~\cite{shu2022test}{\fontsize{4}{12} \selectfont (NeurIPS'22)} & 71.1 & 67.5 & 70.9 & 57.9 & 65.6 & 39.7 \\

& TDA~\cite{karmanov2024efficient}{\fontsize{4}{12} \selectfont (CVPR'24)} & 71.4 & 68.2 & 69.7 & 49.9 & 65.2 & 39.9 \\

& DPE~\cite{zhang2024dual}{\fontsize{4}{12} \selectfont (NeurIPS'24)} & 73.1 & 69.6 & 71.3 & 54.2 & 66.7 & 39.4 \\

& PromptAlign~\cite{abdul2023align}{\fontsize{4}{12} \selectfont (NeurIPS'23)} & 75.3 & 71.6 & 74.6 & 53.2 & 67.1 & 39.7 \\

& ReTA~\cite{liang2025advancing}{\fontsize{4}{12} \selectfont (ACMMM'25)} & 75.1 & 71.3 & 71.8 & 52.8 & 67.6 & 39.8 \\

& MCP~\cite{chen2025multi}{\fontsize{4}{12} \selectfont (ICCV'25)} & 75.7 & 73.0 & 75.5 & 58.2 & 67.6 & 40.1 \\

& T3AL~\cite{liberatori2024test}{\fontsize{4}{12} \selectfont (CVPR'24)} & 76.1 & 72.9 & 75.9 & 59.4 & 67.9 & 40.7 \\

& CRG~\cite{zhai2025mitigating}{\fontsize{4}{12} \selectfont (ICLR'25)}& 76.4 & 73.8 & 76.5 & 60.3 & 68.1 & 40.8 \\

\rowcolor{pinkNonTTA}
\cellcolor{white}\multirow{-7}{*}{FT\,+\,TTA} & {\ours w/o Personalized Cache} & 78.1 & 77.9 & 79.2 & 65.8 & 68.8 & 41.0 \\

\rowcolor{pinkTTA}
\cellcolor{white}
 & {\textbf{\ours (ours)}} & \textbf{81.0} & \textbf{78.6} & \textbf{81.5} & \textbf{67.9} & \textbf{69.2} & \textbf{41.1} \\
\bottomrule
\end{tabular}
}
\caption{Performance (\fonescore score and \war) of zero-shot (ZS), fine-tuning (FT), and TTA on CLIP models for FER. Results are averaged over 10 target subjects for each dataset. CLIP-ViT-B/32\textsuperscript{$\dagger$} denotes full CLIP fine-tuning.}
\label{tab:source_vs_tta_results}
\end{table*}

\begin{table*}[t!]
\centering
\scriptsize

\resizebox{0.9\linewidth}{!}{%
\begin{tabular}{l|cccccccccc|c}
\toprule
\textbf{Method} & Sub-1 & Sub-2 & Sub-3 & Sub-4 & Sub-5 & Sub-6 & Sub-7 & Sub-8 & Sub-9 & Sub-10 & Avg. \\
\midrule
TPT~\cite{shu2022test}{\fontsize{4}{12} \selectfont (NeurIPS'22)}          & 92.0 & 51.5 & 40.0 & 49.6 & 79.0 & 87.0 & 79.5 & 43.4 & 99.0 & 54.0 & 67.5 \\
TDA~\cite{karmanov2024efficient}{\fontsize{4}{12} \selectfont (CVPR'24)}           & 94.9 & 50.3 & 43.5 & 48.0 & 80.8 & 86.2 & 80.2 & 43.5 & \textbf{100.0} & 55.0 & 68.2 \\
DPE~\cite{zhang2024dual}{\fontsize{4}{12} \selectfont (NeurIPS'24)}          & 90.4 & 51.5 & 43.5 & 73.1 & 72.0 & 83.6 & 79.8 & 48.5 & \textbf{100.0} & 54.4 & 69.6 \\
PromptAlign~\cite{abdul2023align}{\fontsize{4}{12} \selectfont (NeurIPS'23)}  & 94.2 & 60.1 & 42.5 & 61.5 & 83.5 & 90.5 & 80.2 & 43.5 & \textbf{100.0} & 60.0 & 71.6 \\
ReTA~\cite{liang2025advancing}{\fontsize{4}{12} \selectfont (ACMMM'25)}         & 94.2 & 60.1 & 43.5 & 60.8 & 83.5 & 88.0 & 80.2 & 43.5 & \textbf{100.0} & 60.0 & 71.3 \\
MCP~\cite{chen2025multi}{\fontsize{4}{12} \selectfont (ICCV'25)}  & 94.5 & 62.0 & 43.5 & 64.0 & 86.0 & 92.0 & 82.0 & 44.0 & \textbf{100.0} & 62.0 & 73.0 \\
T3AL~\cite{liberatori2024test}{\fontsize{4}{12} \selectfont (CVPR'24)}        & \textbf{95.8} & 62.0 & \textbf{43.9} & 61.5 & 83.9 & 93.9 & 83.0 & 43.5 & \textbf{100.0} & 62.0 & 72.9 \\
CRG~\cite{zhai2025mitigating}{\fontsize{4}{12} \selectfont (ICLR'25)}         & 95.0 & 63.0 & 43.5 & 66.0 & 87.0 & 93.0 & 83.0 & 44.5 & \textbf{100.0} & 63.0 & 73.8 \\
\rowcolor{pinkNonTTA}
\ours w/o Personalized Cache
& 94.0 & 65.6 & 32.6 & 72.5 & 99.3
& 96.7 & 89.1 & 47.1 & 99.3 & 83.2
& 77.9 \\
\rowcolor{pinkTTA}
\textbf{\ours (ours)} & 94.3 & \textbf{66.3} & 33.3 & \textbf{73.2} & \textbf{100.0} & \textbf{97.4} & \textbf{89.8} & \textbf{47.8} & \textbf{100.0} & \textbf{84.0} & \textbf{78.6} \\

\bottomrule
\end{tabular}%
 }
\caption{\fonescore score per subject on the \biovid dataset for \ours and competing TTA methods. Best results are shown in \textbf{bold}.}
\label{tab:biovid_subject_full_metrics_f1}
\end{table*}

\begin{table*}[t!]
\centering
\scriptsize

\vspace{0.3em}
\resizebox{0.9\linewidth}{!}{%
\begin{tabular}{l|cccccccccc|c}
\toprule
\textbf{Method} & Sub-1 & Sub-2 & Sub-3 & Sub-4 & Sub-5 & Sub-6 & Sub-7 & Sub-8 & Sub-9 & Sub-10 & Avg. \\
\midrule
TPT~\cite{shu2022test}{\fontsize{4}{12} \selectfont (NeurIPS'22)}          & 70.7 & 43.6 & 86.0 & \textbf{83.1} & 36.4 & 78.7 & 41.1 & 50.0 & 40.0 & 50.0 & 57.9 \\
TDA~\cite{karmanov2024efficient}{\fontsize{4}{12} \selectfont (CVPR'24)}          & 45.5 & 40.0 & 50.1 & 81.0 & 35.0 & 80.0 & 29.6 & 45.6 & 42.9 & 50.0 & 49.9 \\
DPE~\cite{zhang2024dual}{\fontsize{4}{12} \selectfont (NeurIPS'24)}          & 41.0 & 40.6 & 49.0 & 81.0 & 49.8 & 80.0 & 56.0 & 46.0 & 42.9 & 56.6 & 54.2 \\
PromptAlign~\cite{abdul2023align}{\fontsize{4}{12} \selectfont (NeurIPS'23)}  & 41.3 & 45.0 & 45.6 & 82.3 & 48.1 & 80.0 & 44.0 & 58.0 & 42.9 & 45.5 & 53.2 \\
ReTA~\cite{liang2025advancing}{\fontsize{4}{12} \selectfont (ACMMM'25)}         & 43.2 & 41.0 & 53.0 & 80.9 & 36.1 & 80.0 & 39.1 & 53.3 & 42.9 & 59.3 & 52.8 \\
MCP~\cite{chen2025multi}{\fontsize{4}{12} \selectfont (ICCV'25)}         & 55.0 & 44.5 & 60.0 & 78.3 & 43.6 & 79.1 & 52.2 & 61.2 & 44.0 & 65.0 & 58.2 \\
T3AL~\cite{liberatori2024test}{\fontsize{4}{12} \selectfont (CVPR'24)}         & 46.3 & 48.6 & 48.6 & 80.1 & \textbf{51.0} & 80.0 & 61.8 & 68.0 & 44.0 & 66.0 & 59.4 \\
CRG~\cite{zhai2025mitigating}{\fontsize{4}{12} \selectfont (ICLR'25)}         & 58.3 & 46.2 & 63.5 & 81.0 & 47.0 & 81.0 & 58.0 & 65.0 & 44.0 & 59.0 & 60.3 \\
\rowcolor{pinkNonTTA}
\ours w/o Personalized Cache
& 75.4 & 48.8 & 81.0 & 82.0 & 43.1 & 85.1 & 61.5 & 68.0 & 45.9 & 67.6 & 65.8 \\
\rowcolor{pinkTTA}
\textbf{\ours (ours)} & \textbf{77.4} & \textbf{50.0} & \textbf{89.1} & 82.0 & 44.4 & \textbf{87.8} & \textbf{62.0} & \textbf{68.8} & \textbf{47.0} & \textbf{70.8} & \textbf{67.9} \\

\bottomrule
\end{tabular}%
 }
\caption{\fonescore score per subject on \stressid for \ours and other TTA methods.}
\label{tab:stressid_subject_full_metrics_f1}
\end{table*}

\section{Results and Discussion}
\label{sec:results}

\subsection{Experimental Protocol}


\noindent\textbf{Datasets.}  
Experiments are conducted on three challenging public video datasets: \biovid~\cite{walter2013biovid}, \stressid~\cite{chaptoukaev2023stressid}, and \bah~\cite{gonzalez2025bah}. Following prior work~\cite{zeeshan2024subject, sharafi2025disentangled, zeeshan2025musaco, zeeshan2025progressive}, cross-subject splits are used to define source and target sets. The \biovid Heat Pain Database (Part A) contains controlled heat-pain recordings covering multiple intensity levels and a neutral condition, together with synchronized facial videos; 77 subjects are assigned to the source set and 10 to the target set. \stressid contains facial videos collected during stress-inducing tasks, together with self-reported affective ratings; 44 subjects are used as source subjects and 10 as target subjects. \bah is a large-scale multimodal dataset for ambivalence and hesitancy recognition, collected through webcams with frame-level annotations; 143 subjects are assigned to the source set and 10 to the target set.

\noindent \textbf{Implementation Details.}
All experiments were conducted on a single NVIDIA A100 GPU (48 GB), using CLIP ViT-B/32 as the backbone. To construct the personalized static cache, the top-$m{=}3$ closest source subjects are selected for each target subject using the diagonal Fr\'echet distance between the mean and diagonal variance of their embeddings. Target statistics are computed from all available unlabeled target data. The dynamic adaptation stage uses a temporal window length $\mathcal{W}{=}3$ and a prototype-margin threshold $\tau_{\Delta}{=}0.05$. The positive cache capacity is set to $5$ with $\tau_h^{+}{=}0.5$, while the negative cache capacity is set to $4$ with $\tau_h^{-}{=}0.8$.

\noindent \textbf{Baseline Methods.}
Following the baseline CLIP-based FER and TTA methods~\cite{foteinopoulou2024emoclip, karmanov2024efficient, abdul2023align}, all results are reported using two metrics: weighted average recall (\war), which corresponds to overall classification accuracy, and \fonescore score (macro-\fonescore), which balances precision and recall under class imbalance.
Baseline comparisons cover both CLIP-based FER approaches and different TTA methods.
For CLIP-based FER, \ours is evaluated against the frozen CLIP-ViT-B/32 backbone (zero-shot), CLIP-ViT-B/32\textsuperscript{$\dagger$}~\cite{foteinopoulou2024emoclip} (CLIP fine-tuned on source), and EmoCLIP ~\cite{foteinopoulou2024emoclip} (adapter-only training). Additional baselines include X-CLIP~\cite{ni2022expanding} and Exp-CLIP~\cite{zhao2025enhancing}, which incorporate trainable projection modules and, for Exp-CLIP, LLM-derived facial-expression descriptions as supervision.
For TTA, \ours is compared against representative methods originally proposed for image recognition, including TPT~\cite{shu2022test}, TDA~\cite{karmanov2024efficient}, DPE~\cite{zhang2024dual}, PromptAlign~\cite{abdul2023align}, and ReTA~\cite{liang2025advancing}, MCP~\cite{chen2025multi}, and CRG~\cite{zhai2025mitigating}, together with an adaptation of the video-based action recognition method T3AL~\cite{liberatori2024test}. All baselines use the same backbone, prompts, source fine-tuning, target-subject access, subject boundaries, and cache-reset protocol. For all TTA baselines and \ours, the prompt template \emph{``a person with an expression of [CLS]''} is used, following prior CLIP-based FER work~\cite{zhao2025enhancing}. Details on baseline methods and prompt selection are provided in the supplementary materials.

\subsection{Comparison with State-of-the-Art Methods}
\label{sec:sota_comparison}

Tab.~\ref{tab:source_vs_tta_results} compares CLIP-based FER baselines, fine-tuning variants, and state-of-the-art TTA methods on \biovid, \stressid, and \bah datasets. Across all datasets, moving from CLIP inference to adaptation (either via fine-tuning or test-time updates) generally improves recognition under personalized expression styles and domain shift. In particular, TTA methods tend to yield larger gains when target subjects deviate from source distributions, highlighting the importance of test-time personalization beyond a fixed encoder. TTA methods differ in how they trade off accuracy for test-time cost. Prompt-based optimization can be effective but typically incurs higher runtime and memory overhead, whereas cache-based adaptation improves efficiency by reusing confident target evidence. In this context, \ours achieves strong accuracy while remaining lightweight: it attains 81.0/78.6 (\war/\fonescore) on \biovid, 81.5/67.9 on \stressid, and 69.2/41.1 on \bah, while preserving a favorable runtime and memory. Results suggest that the gains are not solely driven by frequent classes (\war), but also translate to improved class-balanced performance (\fonescore score), which is critical under imbalanced labels and heterogeneous subject expression styles.

\noindent\textbf{Complementary Contributions of Tri-Gated Dynamic Adaptation and Static Personalization.}
To determine whether the performance gains arise primarily from the personalized static cache, Tab.~\ref{tab:source_vs_tta_results} compares \textit{TTA-CaP w/o Personalized Cache} with \textbf{TTA-CaP (ours)}. Using only the tri-gate-controlled positive and negative target caches, \textbf{TTA-CaP w/o Personalized Cache} achieves 78.1/77.9, 79.2/65.8, and 68.8/41.0 \war/\fonescore on \biovid, \stressid, and \bah, respectively, outperforming all competing methods. Adding the personalized static cache in \textbf{TTA-CaP (ours)} further improves performance to 81.0/78.6, 81.5/67.9, and 69.2/41.1. These results show that the gains are not solely due to the personalized static cache; instead, the tri-gated dynamic caches and static subject personalization provide complementary improvements.

\begin{table}[t!]
\centering
\footnotesize
\begin{tabular}{lccc}
\toprule
Personalization protocol & \war (\%) & \fonescore (\%) & Time \\
\midrule
Subject-level & \textbf{81.0} & \textbf{78.6} & $\textbf{1.0}\times$ \\
Video-level (online)  & 80.8 & 78.0 & $\sim4.0\times$ \\
\bottomrule
\end{tabular}
\caption{Comparison of subject-level and video-level personalization for \ours on \biovid.}
\label{tab:video_level}
\end{table}

\begin{figure}[t!]
\centering

\begin{tikzpicture}
\begin{axis}[
    width=0.98\linewidth,
    height=0.65\linewidth,
    xlabel={Number of selected source subjects ($m$)},
    ylabel={\war\ (\%)},
    xmin=-0.15,
    xmax=5.15,
    ymin=62,
    ymax=85,
    xtick={0,1,2,3,4,5},
    ytick={65,70,75,80,85},
    tick label style={font=\scriptsize},
    label style={font=\scriptsize},
    axis line style={black!70},
    tick style={black!70},
    grid=major,
    xmajorgrids=false,
    ymajorgrids=true,
    major grid style={
        line width=0.25pt,
        draw=gray!30
    },
    legend style={
        at={(0.5,1.16)},
        anchor=north,
        legend columns=3,
        draw=none,
        fill=white,
        fill opacity=0.85,
        text opacity=1,
        font=\scriptsize,
        column sep=3pt
    },
    clip=false
]

\addplot[
    gray!65,
    dashed,
    line width=0.8pt,
    forget plot
] coordinates {
    (3,62)
    (3,85)
};

\node[
    font=\scriptsize,
    text=gray!70,
    anchor=south
] at (axis cs:3,82.5) {$m=3$};

\addplot[
    biovidcolor,
    line width=1.2pt,
    mark=*,
    mark size=2.2pt
] coordinates {
    (0,71.4)
    (1,76.2)
    (2,78.2)
    (3,81.0)
    (4,81.0)
    (5,81.0)
};
\addlegendentry{\biovid}

\addplot[
    stresscolor,
    line width=1.2pt,
    mark=square*,
    mark size=2.1pt
] coordinates {
    (0,69.9)
    (1,77.6)
    (2,81.5)
    (3,81.5)
    (4,81.8)
    (5,81.5)
};
\addlegendentry{\stressid}

\addplot[
    bahcolor,
    line width=1.2pt,
    mark=triangle*,
    mark size=2.5pt
] coordinates {
    (0,63.8)
    (1,66.9)
    (2,68.4)
    (3,69.2)
    (4,69.2)
    (5,69.2)
};
\addlegendentry{\bah}

\end{axis}
\end{tikzpicture}

{\scriptsize\textbf{(a) Effect of personalized cache size.}}

\vspace{3mm}
\includegraphics[
    width=0.9\linewidth
]{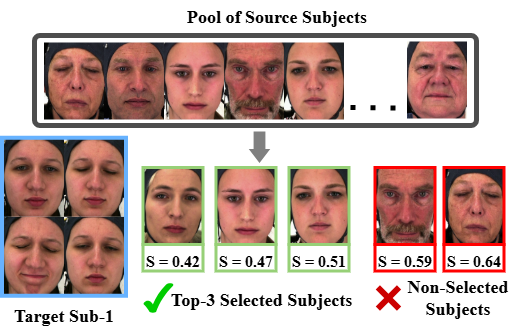}

\vspace{1mm}
{\scriptsize\textbf{(b) Personalized source-subject matching.}}

\caption{\textbf{(a)} Effect of number of selected source subjects on \war. \textbf{(b)} Top-$3$ source subjects selected by diagonal Fr\'echet distance.}
\label{fig:personalized_cache_selection}
\end{figure}

The subject-wise results (Tabs.~\ref{tab:biovid_subject_full_metrics_f1} and \ref{tab:stressid_subject_full_metrics_f1}) further clarify these trends. \textit{TTA-CaP w/o Personalized Cache} also remains competitive across subjects, confirming that the gains are not solely attributable to the static cache. On \biovid and \stressid, personalized test-time evidence generally improves performance across challenging target subjects, although some subject-level variability remains. A paired two-sided Wilcoxon signed-rank test on subject-wise \war across the 10 \biovid target subjects confirms that \ours significantly outperforms T3AL ($p=0.039<0.05$). On \bah, performance is generally lower, and the margins between methods are smaller, likely due to less controlled face-capture conditions that amplify appearance and motion variability. Overall, these results suggest that personalization should be assessed not only through aggregate metrics but also through subject-level performance, and that \ours provides a strong accuracy--efficiency trade-off among CLIP-based TTA approaches. Due to space constraints, subject-wise \fonescore for the 10 target subjects on \biovid and \stressid is reported in the main paper; complete subject-wise results for \bah, subject-wise \war results for all datasets, and subject-wise results for the CLIP-based FER baselines are provided in the supplementary materials.

\subsection{Ablation Studies}
\label{sec:ablation}

\noindent\textbf{Ablation on Online Personalization.} To evaluate a stricter online setting, we conduct an ablation on \biovid where personalization is performed independently for each target video. The personalized static cache is constructed using only the current video, without other unlabeled videos from the same subject. As shown in Tab.~\ref{tab:video_level}, video-level personalization performs comparably to the default subject-level setting, with only a 0.2\% decrease in \war and a 0.6\% decrease in \fonescore. This suggests that a single video provides sufficient information for effective personalization. However, it requires approximately $4\times$ more processing time because the static cache must be reconstructed for each video rather than reused across videos of the same subject.


\noindent\textbf{Impact of Personalized Source-Subject Selection.}
Fig.~\ref{fig:personalized_cache_selection} (a) shows that selecting a few matched subjects for personalized static cache improves \war across all datasets. Performance largely saturates at $m=2$--$3$, suggesting that a small set of closely aligned subjects provides sufficient reference information. Accordingly, $m=3$ is used in all subsequent experiments. Fig.~\ref{fig:personalized_cache_selection} (b) illustrates the matching process, where each target subject is compared with all 77 source subjects using the diagonal Fr'echet distance, and the top-$m$ subjects are selected to construct the personalized static cache.

\noindent\textbf{Component Analysis.}
We further conduct a component-wise ablation on BioVid to quantify the contribution of the main components of \ours. Starting from the full method, we individually remove the positive cache, negative cache, tri-gate mechanism, temporal encoder, personalized static cache, and embedding-level fusion. Results are reported in Tab.~\ref{tab:component_ablation}. The component analysis shows how each part of the framework contributes to the final performance and provides a more complete separation between the gains from temporal modeling, personalization, cache-based adaptation, gating, and embedding-level fusion.

\begin{table}[t]
\centering
\small
\resizebox{0.8\linewidth}{!}{%
\begin{tabular}{lcc}
\toprule
Method & \war (\%) & \fonescore (\%) \\
\midrule
\ours & \textbf{81.0} & \textbf{78.6} \\
w/o Positive Cache & 79.2 & 76.8 \\
w/o Negative Cache & 80.0 & 77.6 \\
w/o Tri-Gate & 78.7 & 76.0 \\
w/o Temporal Encoder & 80.6 & 78.1 \\
w/o Personalized Static Cache & 78.1 & 77.9 \\
w/o Embedding-Level Fusion & 79.0 & 76.7 \\
\bottomrule
\end{tabular}
}
\caption{Component ablation of \ours on \biovid. Each row removes one component from the full method.}
\label{tab:component_ablation}
\end{table}

\noindent\textbf{Ablation on Cache-Update Gating.}
Fig.~\ref{fig:gate_loo} reports a leave-one-out ablation of the tri-gate mechanism on \biovid. Removing the temporal, entropy, and prototype gates reduces \war from 81.0\% to 79.2\%, 79.7\%, and 78.5\%, respectively. The largest degradation occurs without the prototype gate, while the full tri-gate achieves the best performance, confirming the complementary contribution of all three gates.

\begin{figure}[t!]
\centering
\begin{tikzpicture}

\begin{axis}[
    width=0.88\linewidth,
    height=0.58\linewidth,
    ybar,
    bar width=16pt,
    ymin=77,
    ymax=82,
    ylabel={\war\ (\%)},
    symbolic x coords={w/o Temporal,w/o Entropy,w/o Prototype,Full},
    xtick=data,
    tick label style={font=\scriptsize},
    label style={font=\small},
    enlarge x limits=0.18,
    grid=major,
    major grid style={
        line width=0.2pt,
        draw=gray!35
    },
    nodes near coords,
    nodes near coords style={
        font=\scriptsize,
        anchor=south,
        /pgf/number format/fixed,
        /pgf/number format/precision=1
    },
    major tick length=0pt,
]

\addplot[
    draw=none,
    fill=teal!35
] coordinates {
    (w/o Temporal,79.2)
    (w/o Entropy,79.7)
    (w/o Prototype,78.5)
    (Full,81.0)
};

\end{axis}

\end{tikzpicture}

\caption{
Leave-one-out ablation of the tri-gate mechanism on \biovid. Removing any gate reduces \war, with the largest drop observed when removing the prototype gate.
}
\label{fig:gate_loo}
\end{figure}

\section{Conclusion}
\label{sec:conclusion}

This paper proposes \ours, a gradient-free cache-based TTA method for cost-effective personalization of CLIP-based video FER.  \ours combines two target dynamic caches with a fixed personalized source cache built from source-subject prototypes to improve subject-specific adaptation at test time. Moreover, a tri-gate reliability mechanism is introduced to update the target caches and refine predictions via feature-level fusion with retrieved cache entries. Experiments on the \biovid, \stressid, and \bah datasets show consistent gains in both \war and \fonescore over prompt-tuning and cache-based baselines, while maintaining low test-time computation and bounded memory usage. These findings support the central premise that subject-specific prototypes provide valuable prior guidance to improve pseudo-label reliability and enable effective personalization in video-based FER. A current limitation is that personalized-cache construction assumes access to target-subject data, preventing fully online deployment. Future work will address this limitation and extend \ours to other video understanding tasks, including action, gesture, event, and affect recognition.\\




\noindent\textbf{Supplementary materials.}
Includes additional subject-wise \fonescore and \war results for TTA methods, baseline and implementation details, algorithms for personalized cache and TTA, and extended ablation studies.

{
    \small
    \bibliographystyle{ieeenat_fullname}
    \bibliography{main}
}

\end{document}


\maketitle


\begingroup
\let\clearpage\relax
\let\newpage\relax

\maketitle
\vspace{-0.5cm}

\renewcommand{\contentsname}{Table of Contents}

\makeatletter
\let\origcontentsline\contentsline
\def\contentsline#1#2#3#4{%
  \ifnum\pdfstrcmp{#1}{title}=0
  \else
    \ifnum\pdfstrcmp{#1}{author}=0
    \else
      \origcontentsline{#1}{#2}{#3}{#4}%
    \fi
  \fi
}
\tableofcontents
\makeatother

\endgroup

\section{Algorithmic Details}
This section provides additional algorithmic details for the personalized source-prototype construction and the test-time adaptation pipeline introduced in the main paper. In particular, we clarify how personalized source prototypes are built from the source subjects and subsequently matched to target subjects, as well as how the proposed adaptation mechanism operates at inference time on target videos.

Algorithm~\ref{alg:personalized_cache} describes the construction of the personalized source prototypes. Representative prototypes are extracted independently for each source subject and expression class using DBSCAN. For each subject--class subset, $\mathrm{min\_samples}\in{5,10,15}$ is evaluated, while candidate neighborhood radii $\epsilon$ are obtained from the ${0.5,0.6,0.7,0.8,0.9}$ quantiles of the corresponding k-NN distance distribution. Degenerate clusterings are discarded, and each remaining configuration is evaluated using a score combining clustering validity and stability, where validity favors a low noise ratio and non-degenerate clusters, and stability is measured by the mean adjusted Rand index (ARI) over six bootstrap runs. The configuration with the highest score is selected, and each non-noise cluster is represented by the observed embedding closest to its centroid. Source prototype extraction, including feature extraction, DBSCAN parameter selection, and clustering, is performed only once before deployment and requires approximately 17.9 minutes. For \biovid, constructing the personalized caches for all 12 target subjects requires an additional 1.38 minutes, or approximately 6.9 seconds per subject. Thus, DBSCAN introduces no repeated clustering cost during test-time adaptation.

The resulting source prototypes are transferred to each target subject through the statistics-based matching strategy described in the main paper. Algorithm~\ref{alg:online_tta} summarizes the subsequent test-time adaptation procedure, including temporal representation extraction, cache retrieval, embedding-level fusion, and tri-gate cache updates. No backbone parameters are updated during adaptation.

\begin{algorithm}[t!]
\caption{Personalized static cache}
\label{alg:personalized_cache}
\small
\KwIn{Source set $\mathcal{D}^{\mathrm{src}}$, target set $\mathcal{D}^{\mathrm{tgt}}$, frozen image encoder $E_I$, classes $\mathcal{C}$, DBSCAN candidate sets $\mathcal{E}$ and $\mathcal{Q}$, top-$m$, cap $K$}
\KwOut{Personalized prototypes $\mathcal{P}^{\mathrm{pers}}_{t,c}$}

\ForEach{source subject $s$ and class $c \in \mathcal{C}$}{
Extract normalized embeddings $\mathbf{F}_{s,c}$;
$\mathcal{R} \leftarrow \emptyset$;

```
\ForEach{$(\epsilon,q)\in\mathcal{E}\times\mathcal{Q}$}{
    Run DBSCAN$(\mathbf{F}_{s,c};\epsilon,q)$\;
    Compute the outlier ratio and number of non-noise clusters\;

    \If{the clustering is non-degenerate}{
        Estimate clustering stability by bootstrap resampling\;
        Compute the mean ARI across bootstrap runs\;
        Add $(\epsilon,q,\text{outlier ratio},\text{ARI})$ to $\mathcal{R}$\;
    }
}

\eIf{$\mathcal{R}=\emptyset$ or too few samples are available}{
    Set $\mathcal{M}_{s,c}$ to the sample nearest to the class mean\;
}{
    Select the candidate with low outlier ratio and highest clustering stability\;
    Run DBSCAN using the selected parameters\;
    Set $\mathcal{M}_{s,c}$ to the observed embedding nearest to each non-noise cluster centroid\;
}
```

}

\ForEach{source subject $s$}{
Compute subject-level statistics $(\mu_s,\tilde{\sigma}_s^2)$
from all source embeddings of subject $s$;
}

\ForEach{target subject $t$}{
Compute target statistics $(\mu_t,\tilde{\sigma}_t^2)$
from all unlabeled target embeddings;

```
\ForEach{source subject $s$}{
    Compute anchor distance $d(s,t)$\;
}

Select the top-$m$ closest sources $\mathcal{S}_t$\;

\ForEach{class $c \in \mathcal{C}$}{
    $\mathcal{P}^{\mathrm{pers}}_{t,c}
    \leftarrow
    \bigcup_{s\in\mathcal{S}_t}\mathcal{M}_{s,c}$\;

    \If{$K>0$ and $|\mathcal{P}^{\mathrm{pers}}_{t,c}|>K$}{
        Keep the $K$ most relevant prototypes\;
    }
}
```

}

\Return{${\mathcal{P}^{\mathrm{pers}}*{t,c}}*{t,c}$};
\end{algorithm}

\begin{algorithm}[t!]
\caption{Test-Time Adaptation}
\label{alg:online_tta}
\small
\DontPrintSemicolon
\SetKwInOut{Input}{Input}
\SetKwInOut{Output}{Output}

\Input{
Target video
$\mathcal{V}^{\mathrm{tgt}}
=\{x_t^{\mathrm{tgt}}\}_{t=1}^{T}$;
frozen image encoder $E_I$;
frozen temporal encoder $E_V$;
frozen text encoder $E_T$;
class prompts $\{\tau_c\}_{c\in\mathcal{C}}$;
personalized static cache
$\mathcal{P}^{\mathrm{src,tgt}}$;
temporal window length $W$;
CLIP logit scale $\eta$.
}

\Output{
Video prediction $\hat{y}$;
final target caches
$\mathcal{C}^{\mathrm{tgt}+}$ and
$\mathcal{C}^{\mathrm{tgt}-}$.
}

Initialize
$\mathcal{C}^{\mathrm{tgt}+}\leftarrow\emptyset$
and
$\mathcal{C}^{\mathrm{tgt}-}\leftarrow\emptyset$\;

\ForEach{$c\in\mathcal{C}$}{
    $e_c
    \leftarrow
    E_T(\tau_c)/
    \lVert E_T(\tau_c)\rVert_2$\;
}

\For{$t=1$ \KwTo $T$}{

    $v_t^{\mathrm{tgt}}
    \leftarrow
    E_I(x_t^{\mathrm{tgt}})$\;

    $\mathcal{I}_t
    \leftarrow
    \{\max(1,t-W+1),\ldots,t\}$\;

    $z_t^{\mathrm{tgt}}
    \leftarrow
    E_V\!\left(
    \{v_j^{\mathrm{tgt}}\}_{j\in\mathcal{I}_t}
    \right)$\;

    $z_t^{\mathrm{tgt}}
    \leftarrow
    z_t^{\mathrm{tgt}}/
    \lVert z_t^{\mathrm{tgt}}\rVert_2$\;

    \tcp{Initial prediction}

    \ForEach{$c\in\mathcal{C}$}{
        $\ell_t^{\mathrm{base}}(c)
        \leftarrow
        \eta\,
        \cos(z_t^{\mathrm{tgt}},e_c)$\;
    }

    $p_t^{\mathrm{base}}
    \leftarrow
    \operatorname{softmax}
    (\ell_t^{\mathrm{base}})$\;

    $\tilde{y}_t
    \leftarrow
    \arg\max_{c\in\mathcal{C}}
    \ell_t^{\mathrm{base}}(c)$\;

    $H_t
    \leftarrow
    -\sum_{c\in\mathcal{C}}
    p_t^{\mathrm{base}}(c)
    \log p_t^{\mathrm{base}}(c)$\;

    \tcp{Cache retrieval and embedding fusion}

    $z_t^{\mathrm{src,tgt}}
    \leftarrow
    \operatorname{RetrieveSrc}
    \bigl(
    \mathcal{P}^{\mathrm{src,tgt}},
    z_t^{\mathrm{tgt}}
    \bigr)$\;

    $z_t^{\mathrm{tgt}+}
    \leftarrow
    \operatorname{RetrievePos}
    \bigl(
    \mathcal{C}^{\mathrm{tgt}+},
    z_t^{\mathrm{tgt}}
    \bigr)$\;

    $z_t^{\mathrm{tgt}-}
    \leftarrow
    \operatorname{RetrieveNeg}
    \bigl(
    \mathcal{C}^{\mathrm{tgt}-},
    z_t^{\mathrm{tgt}}
    \bigr)$\;

    $z_t^{\mathrm{fuse}}
    \leftarrow
    z_t^{\mathrm{tgt}}
    +z_t^{\mathrm{src,tgt}}
    +z_t^{\mathrm{tgt}+}
    -z_t^{\mathrm{tgt}-}$\;

    $z_t^{\mathrm{fuse}}
    \leftarrow
    z_t^{\mathrm{fuse}}/
    \lVert z_t^{\mathrm{fuse}}\rVert_2$\;

    \ForEach{$c\in\mathcal{C}$}{
        $\ell_t(c)
        \leftarrow
        \eta\,
        \cos(z_t^{\mathrm{fuse}},e_c)$\;
    }

    \tcp{Update caches after retrieval}

    $\mathcal{H}_t
    \leftarrow
    \{\tilde{y}_j\}_{j\in\mathcal{I}_t}$\;

    $\begin{aligned}
    \bigl(
    \mathcal{C}^{\mathrm{tgt}+},
    \mathcal{C}^{\mathrm{tgt}-}
    \bigr)
    \leftarrow
    \operatorname{TriGateUpdate}\bigl(
    &z_t^{\mathrm{tgt}},
    \tilde{y}_t,
    H_t,
    \mathcal{H}_t,\\[-0.2em]
    &\mathcal{P}^{\mathrm{src,tgt}},
    \mathcal{C}^{\mathrm{tgt}+},
    \mathcal{C}^{\mathrm{tgt}-}
    \bigr)
    \end{aligned}$\;
}

\ForEach{$c\in\mathcal{C}$}{
    $\bar{\ell}(c)
    \leftarrow
    \frac{1}{T}
    \sum_{t=1}^{T}\ell_t(c)$\;
}

$\hat{y}
\leftarrow
\arg\max_{c\in\mathcal{C}}
\bar{\ell}(c)$\;

\KwRet{
$\hat{y},
\mathcal{C}^{\mathrm{tgt}+},
\mathcal{C}^{\mathrm{tgt}-}$
}\;
\end{algorithm}

\section{TTA Baseline Implementation Details}
We compare against several CLIP-based test-time adaptation baselines. \emph{TPT}~\cite{shu2022test} updates only the textual prompt tokens during inference by minimizing prediction entropy over strongly augmented views, while keeping both the image and text encoders frozen. \emph{TDA}~\cite{karmanov2024efficient} is a training-free dynamic adapter that builds a feature--label cache online, progressively refines pseudo-labels, including negative pseudo-labels, and combines cache-based predictions with CLIP zero-shot logits. \emph{DPE}~\cite{zhang2024dual} updates visual and textual class prototypes jointly at test time, introduces lightweight sample-specific residuals, and encourages cross-modal consistency between the two prototype spaces under distribution shift. \emph{PromptAlign}~\cite{abdul2023align} adapts prompts by aligning test-time feature statistics with source-domain statistics, thereby reducing distribution mismatch beyond standard entropy-minimization objectives. \emph{ReTA}~\cite{liang2025advancing} emphasizes reliable cache-based VLM adaptation through CER (consistency-aware entropy reweighting) for selective cache updates and DDC (diversity-driven distribution calibration), which models each class as a Gaussian family over evolving text embeddings to improve calibration. Finally, \emph{T3AL}~\cite{liberatori2024test}, originally proposed for zero-shot temporal action localization, is adapted here for classification by using video-level pseudo-labels from aggregated frame features, self-supervised refinement of frame-level predictions, and text-guided suppression of irrelevant regions, while keeping the encoders frozen and resetting updates for each sample. In addition, we include \emph{MCP}~\cite{chen2025multi}, which performs test-time adaptation by refining class predictions through multi-modal prototype-based calibration, leveraging the complementary information encoded in visual and textual representations. We also compare with \emph{CRG}~\cite{zhai2025mitigating}, which improves test-time predictions by exploiting cross-modal relationships to refine the representation and prediction space, providing a training-free mechanism for adapting CLIP to the target distribution.

\section{Source Training and Temporal Encoder Details}
\label{sec}

The source model uses CLIP with a ViT-B/32 image encoder and a temporal encoder based on the Transformer-based temporal modeling approach of EmoCLIP~\cite{foteinopoulou2024emoclip}. Frame-level CLIP representations are passed to a four-layer Transformer encoder with eight attention heads, a feed-forward dimension of 2048, dropout of 0.1, and an embedding dimension of 512. A learnable classification token aggregates the temporal information, and its output is used as the video representation. The model operates on 8-frame clips.

The temporal encoder is initialized from scratch and trained on the labeled source data together with the CLIP backbone. Training uses a contrastive objective that increases the similarity between the video representation and the text embedding of its corresponding class while decreasing similarity to the other classes. Training is performed for 30 epochs using AdamW with a learning rate of $10^{-4}$. The same source-trained ViT-B/32 backbone, temporal encoder, frame windows, and source checkpoint are used for all TTA baselines and \ours.

\section{Computation Complexity}
\label{sec:complexity}
Tab.~\ref{tab:runtime_biovid} reports the per-batch runtime (ms), memory (MB), and average \war (\%) on the \biovid dataset.  Overall, \ours achieves the best accuracy while keeping the test-time cost low.  In particular, \ours is faster than prompt-tuning TTA methods (e.g., TPT and PromptAlign), since it avoids backpropagation and instead performs a small number of encoder forward passes, cache retrievals, and a lightweight fusion step. 
As a result, \ours offers a markedly improved accuracy--efficiency trade-off over existing TTA baselines, improving \war while maintaining low runtime and memory.

\section{Additional Ablation Studies}

\subsection{Effect of Negative-Cache Subtraction.}
To examine the effect of the negative cache, we compute class scores as the cosine similarity between target representations and class-text embeddings, before and after subtracting the retrieved negative representation on \stressid. Tab.~\ref{tab:negative_cache_scores} reports these scores averaged over all target samples for each pseudo-label. The non-current score is defined as the highest score among classes other than the current pseudo-label. Ideally, negative-cache subtraction should suppress the non-current score more strongly than the current-class score, thereby increasing the class margin.

As shown in Tab.~\ref{tab:negative_cache_scores}, negative-cache subtraction consistently increases the average class margin. For samples predicted as \textit{stress}, the margin increases from 0.14 to 0.25, as the non-current score decreases more strongly (0.47 to 0.33) than the current-class score (0.61 to 0.58). Similarly, for \textit{neutral}, the margin increases from 0.18 to 0.29. These results show that the negative cache suppresses non-current-class evidence while largely preserving the current-class score.

\begin{table}[t!]
\centering
\scriptsize
\setlength{\tabcolsep}{6pt}
\begin{tabular}{l|ccr}
\toprule
\textbf{Method} & \textbf{Time (ms)} & \textbf{Memory (MB)} & \multicolumn{1}{r}{\textbf{\war (\%)}}\\
\midrule
TPT & 771.2 & 2800 & 71.1\\
PromptAlign & 900.0 & 3100 & 75.3\\
TDA &  \textbf{97.8} & 2124 & 71.4\\
DPE & 263.9 & 1700 & 73.1\\
MCP   & 145.0 & 1450 & 75.7 \\
ReTA & 174.0 & 2390 & 75.1\\
T3AL & 520.0 & 2200 & 76.1\\
CRG   & 205.0 & 1850 & 76.4 \\
\rowcolor{pinkTTA}
\textbf{\ours} & 110.0 & \textbf{904.0} & \textbf{81.0}\\
\bottomrule
\end{tabular}
\caption{Complexity of TTA methods. Run time per batch (B=16), memory, and average \war for \biovid target subjects.}
\label{tab:runtime_biovid}
\end{table}


\begin{table}[t!]
\centering
\small
\begin{tabular}{lccc}
\toprule
Representation & Current-class & Non-current-class & $\Delta$ \\
\midrule
\multicolumn{4}{c}{\textit{Pseudo-label: Stress}} \\
$\mathbf{z}^{\mathrm{tgt}}$
& 0.61 & 0.47 & 0.14 \\
$\mathbf{z}^{\mathrm{tgt}}-\mathbf{z}^{\mathrm{tgt}-}$
& 0.58 & 0.33 & 0.25 \\
\midrule
\multicolumn{4}{c}{\textit{Pseudo-label: Neutral}} \\
$\mathbf{z}^{\mathrm{tgt}}$
& 0.63 & 0.45 & 0.18 \\
$\mathbf{z}^{\mathrm{tgt}}-\mathbf{z}^{\mathrm{tgt}-}$
& 0.60 & 0.31 & 0.29 \\
\bottomrule
\end{tabular}
\caption{Effect of negative-cache subtraction in \ours on \stressid. Class scores are cosine similarities between target representations and class-text embeddings, averaged over all target samples for each pseudo-label. The non-current score is the highest score among classes other than the current pseudo-label, and $\Delta$ denotes the margin between the current- and highest non-current-class scores.}
\label{tab:negative_cache_scores}
\end{table}

\begin{table}[t!]
\centering
\small
\begin{tabular}{lcc}
\toprule
Method & \war (\%) & \fonescore (\%) \\
\midrule
MCP & 75.7 & 73.0 \\
MCP + Temporal & 75.9 & 73.4 \\
MCP + Temporal + Personalized Cache & 77.6 & 75.3 \\
\ours & \textbf{81.0} & \textbf{78.6} \\
\bottomrule
\end{tabular}
\caption{Fair comparison with MCP on \biovid. Temporal modeling and personalized cache denote the corresponding components used in \ours.}
\label{tab:fair_mcp}
\end{table}

\begin{table}[t!]
\centering

\setlength{\tabcolsep}{5pt}
\begin{tabular}{lcc}
\toprule
Backbone & Method & \war (\%) \\
\midrule
\multirow{3}{*}{RN101}
& Source FT& 64.8 \\
& TTA-CaP  & \textbf{76.9} \\
\midrule
\multirow{3}{*}{ViT-B/16}
& Source FT& 66.1 \\
& TTA-CaP  & \textbf{79.6} \\
\midrule
\multirow{3}{*}{ViT-B/32}
& Source FT& 67.7 \\
& TTA-CaP  & \textbf{81.0} \\
\bottomrule
\end{tabular}
\caption{Performance of \ours with different backbones on \biovid.}
\label{tab:backbone_biovid}
\end{table}

\begin{table}[t!] 
\centering 
\begin{tabular}{cccc} 
\toprule 
$m$ & Time (ms) & Memory (MB) & \war (\%) \\ 
\midrule 
1 & 104 & 870 & 76.2 \\ 2 & 107 & 885 & 78.2 \\ 
3 & \textbf{110} & \textbf{904} & \textbf{81.0} \\ 
4 & 113 & 921 & 81.0 \\ 5 & 116 & 938 & 81.0 \\ 
\bottomrule \end{tabular} 
\caption{Accuracy--efficiency tradeoff of \ours for different personalized cache sizes ($m$) on \biovid.} \label{tab:cache_tradeoff} 
\end{table}

\begin{table}[t!]
\centering
\begin{tabular}{ccc}
\toprule
Window size & \ours w/o PSC & \ours \\
\midrule
16 & 41.1 & 41.0 \\
24 & 42.0 & 43.5 \\
32 & \textbf{42.2} & \textbf{43.8} \\
\bottomrule
\end{tabular}
\caption{Effect of temporal window size on the \fonescore (\%) of \ours on \bah. PSC denotes the personalized static cache.}
\label{tab:window_bah}
\end{table}

\subsection{Fair Comparison with Baseline Methods}
\label{sec:fair_mcp}

To further isolate the contribution of temporal modeling and personalization, we augment MCP with the same temporal modeling and personalized static cache used in \ours and evaluate all variants on \biovid. This comparison progressively introduces these components into MCP while keeping the remaining experimental protocol unchanged. The results are reported in Tab.~\ref{tab:fair_mcp}.

Adding temporal modeling alone yields only a marginal improvement for MCP, from 75.7\% to 75.9\% in \war and from 73.0\% to 73.4\% in \fonescore, suggesting that temporal modeling by itself does not account for the performance gain of \ours. Incorporating the personalized cache provides a more noticeable improvement, raising MCP to 77.6\% \war and 75.3\% \fonescore. Nevertheless, \ours achieves substantially higher performance (81.0\% \war and 78.6\% \fonescore), indicating that its gains cannot be explained solely by the use of temporal modeling or personalization, but rather by the overall adaptation strategy.

\subsection{Backbone Generalization}
\label{sec:backbone_generalization}

To evaluate the robustness of \ours to different visual backbones, additional experiments are conducted on \biovid using RN101, ViT-B/16, and ViT-B/32. For each backbone, the corresponding source fine-tuned model is compared with \ours under the same adaptation protocol. As shown in Tab.~\ref{tab:backbone_biovid}, \ours consistently improves over the corresponding source fine-tuned model for all backbones, with gains of 12.1, 13.5, and 13.3 percentage points in \war for RN101, ViT-B/16, and ViT-B/32, respectively. These results indicate that the proposed adaptation strategy is not specific to a particular CLIP backbone.

\subsection{Accuracy--Efficiency Tradeoff of Personalized Cache Size}
\label{sec:cache_tradeoff}

We evaluate the effect of the personalized cache size $m$ on both performance and computational cost on \biovid. As shown in Tab.~\ref{tab:cache_tradeoff}, increasing $m$ from 1 to 3 improves \war from 76.2\% to 81.0\% with a moderate increase in runtime and memory. Beyond $m=3$, performance saturates while the computational cost continues to increase, supporting $m=3$ as a favorable tradeoff.

\begin{figure}[t!]
  \centering
  \includegraphics[width=0.9\linewidth]{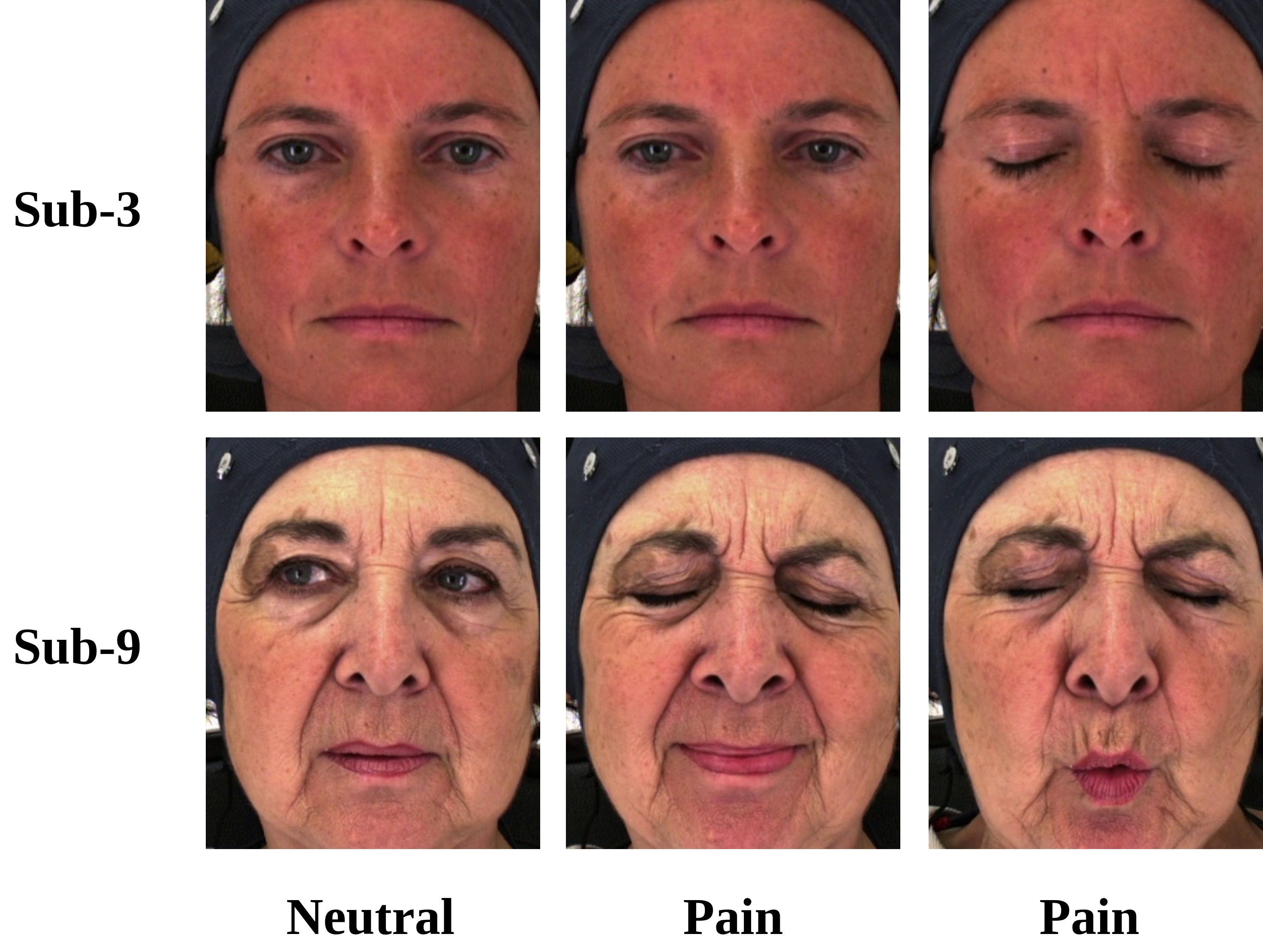}
  \caption{Representative neutral and pain samples from an easier subject (Sub-9) and a more challenging subject (Sub-3). Sub-9 exhibits more visually distinguishable expression changes between neutral and pain, whereas the two classes are less separable for Sub-3, consistent with their subject-wise \fonescore results.}
  \label{fig:sub9}
\end{figure}

\begin{table}[t!]
\centering
\scriptsize
\setlength{\tabcolsep}{8pt}
\begin{tabular}{lcc}
\toprule
\textbf{Method} & \textbf{Logit ECE} $\downarrow$ & \textbf{Embed ECE} $\downarrow$ \\
\midrule
TDA  & 20.7 & 15.2 \\
ReTA & 19.2 & 14.0 \\
\ours  & \textbf{17.8} & \textbf{12.3} \\
\bottomrule
\end{tabular}
\caption{Calibration comparison on \biovid}
\label{tab:ece_calibration}
\end{table}

\begin{table}[t!]
\centering
\setlength{\tabcolsep}{20pt}
\scriptsize
\renewcommand{\arraystretch}{1.15}
\setlength{\tabcolsep}{7pt}
\begin{tabular}{lr}
\toprule
\textbf{Weighting Strategy} & \textbf{\war\ (\%)} $\uparrow$ \\
\midrule
Conf.-weighted  & 78.9 \\
Attention-based & 77.4 \\
Equal-weight    & \textbf{81.0} \\
\bottomrule
\end{tabular}
\caption{Ablation of cache-weighting strategies for \ours on \biovid. Results report the average \war over target subjects.}
\label{tab:fusion_ablation}
\end{table}

\begin{table}[t]
\centering
\small
\setlength{\tabcolsep}{5pt}
\renewcommand{\arraystretch}{1.1}
\begin{tabular}{lcc}
\toprule
\textbf{Prompt Template} & \textbf{\war} & \textbf{\fonescore score} \\
\midrule
a photo of a [CLS]                                      & 68.7 & 63.3 \\
a\_photo\_of\_the\_[CLS]\_face                          & 66.5 & 59.3 \\
a\_photo\_of\_one\_[CLS]\_face                          & 64.2 & 55.8 \\
a\_close-up\_photo\_of\_the\_[CLS]\_face                & 68.2 & 62.6 \\
a\_low\_resolution\_photo\_of\_a\_[CLS]\_face           & 68.7 & 64.2 \\
a\_good\_photo\_of\_a\_[CLS]\_face                      & 69.2 & 65.4 \\
a\_photo\_of\_my\_[CLS]\_face                           & 66.2 & 60.3 \\
a\_cropped\_photo\_of\_the\_[CLS]\_face                 & 64.5 & 57.5 \\
a\_photo\_of\_a\_person\_with\_[CLS]\_face              & 67.2 & 61.4 \\
\rowcolor{pinkTTA}
\textbf{a person with an expression of [CLS]}           & \textbf{69.7} & \textbf{66.6} \\
a\_portrait\_of\_a\_person\_in\_[CLS]                   & 58.2 & 46.7 \\
a\_face\_showing\_signs\_of\_[CLS]                      & 68.7 & 64.1 \\
a close up portrait of a [CLS] expression               & 69.2 & 63.0 \\
a realistic photo of a person experiencing [CLS]        & 53.5 & 40.0 \\
a cropped image of a person in [CLS]                    & 69.0 & 64.5 \\
a photo of a [CLS] person                               & 69.5 & 64.3 \\
a person with a facial expression of [CLS]              & 69.5 & 64.2 \\
a high quality photo of a [CLS] expression              & 69.5 & 64.3 \\
a photo of a face showing [CLS]                         & 69.5 & 64.3 \\
a photo of a face in [CLS]                              & 69.0 & 61.8 \\
\bottomrule
\end{tabular}
\caption{Ablation of class-prompt templates for the CLIP text encoder on \biovid using CLIP ViT-B/32. Performance varies noticeably across prompt formulations, and the template \emph{“a person with an expression of [CLS]”} achieves the best \war and \fonescore score.}
\label{tab:class_prompt_ablation_full}
\end{table}

\begin{table}[t!]
\centering
\small
\setlength{\tabcolsep}{6pt}
\begin{tabular}{lc}
\toprule
\textbf{Cache construction} & \textbf{\war(\%)} $\uparrow$ \\
\midrule
Baseline (k-means, $k{=}1$) & 78.8 \\
Two-stage (k-means) & 79.3 \\
Two-stage (DBSCAN) & 79.5 \\
Personalized (k-means) & 79.6 \\
Personalized (DBSCAN) & \textbf{81.0} \\
\bottomrule
\end{tabular}
\caption{Effect of different static source-cache construction strategies. Average \war{} (\%) over target subjects in \biovid using \ours{}.}
\label{tab:src_cache_build}
\end{table}

\subsection{Analysis of Subject-Level Performance Variation}

To better understand the large variation in subject-wise performance, we qualitatively compare an easier subject (Sub-9) with a more challenging subject (Sub-3). As illustrated in Fig.~\ref{fig:sub9}, Sub-9 exhibits relatively distinct visual appearance between neutral and pain expressions, providing a stronger discriminative signal for classification. This is consistent with the subject-wise results, where most methods achieve an \fonescore close to or equal to 100\% for Sub-9. In contrast, the neutral and pain expressions of Sub-3 are visually less distinguishable, leading to substantially lower performance, with \fonescore values of approximately 40\% for most methods. These observations indicate that the unusually high performance on Sub-9 is associated with subject-specific expression separability and is consistently observed across different methods, rather than being specific to \ours.

\begin{figure}[t!]
\centering
\begin{tikzpicture}
\begin{axis}[
  ybar,
  bar width=14pt,
  bar shift=0pt,
  width=0.85\linewidth,
  height=4.2cm,
  ymin=0, ymax=100,
  enlarge x limits=0.15,
  ylabel={\war{} (\%)},
  xtick={0,1,2,3,4},
  xticklabels={GMM,Random,Classifier,Generator,Prototypes},
  xtick style={draw=none},
  xticklabel style={font=\scriptsize, rotate=15, anchor=north east},
  yticklabel style={font=\scriptsize},
  ylabel style={font=\scriptsize},
  grid=major,
  grid style={dashed, gray!30},
  nodes near coords,
  every node near coord/.append style={
    font=\scriptsize,
    /pgf/number format/fixed,
    /pgf/number format/precision=1
  },
]
\definecolor{barA}{HTML}{C85A6A}
\definecolor{barB}{HTML}{1F7A73}
\definecolor{barD}{HTML}{D8A23A}
\definecolor{barE}{HTML}{4E79A7}
\definecolor{barF}{HTML}{59A14F}

\addplot[fill=barD, fill opacity=0.85, draw=none]
coordinates {(0,74.0)};
\addplot[fill=barF, fill opacity=0.85, draw=none]
coordinates {(1,77.3)};
\addplot[fill=barE, fill opacity=0.85, draw=none]
coordinates {(2,77.6)};
\addplot[fill=barB, fill opacity=0.85, draw=none]
coordinates {(3,79.1)};
\addplot[fill=barA, fill opacity=0.85, draw=none]
coordinates {(4,81.0)};
\end{axis}
\end{tikzpicture}
\caption{Ablation of source-cache construction on \biovid. Results
report the average \war{} across target subjects. ``Prototypes''
denotes the personalized source cache built from source subject
prototypes.}
\label{fig:src_cache_gen_war}
\end{figure}

\begin{figure}[t!]
  \centering
  \includegraphics[width=0.9\linewidth]{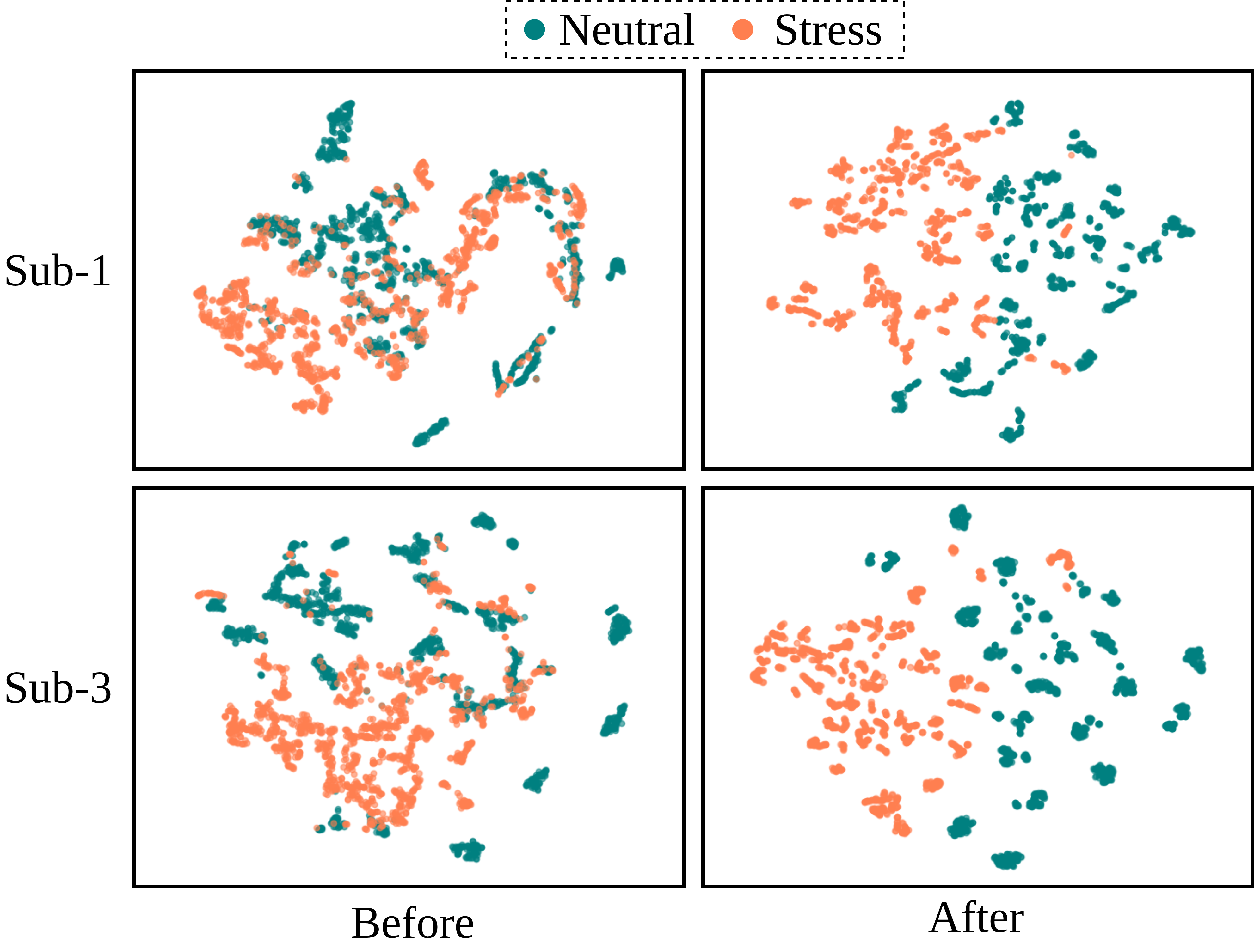}
  \caption{t-SNE visualization of the embedding space \emph{before} and \emph{after} cache fusion for two subjects from the \stressid dataset.
  }
  \label{fig:qual_tsne}
\end{figure}

\subsection{Embedding vs.\ Logit calibration}
Tab.~\ref{tab:ece_calibration} compares the calibration of logit-space and embedding-space predictions on \biovid. Embedding-space fusion consistently yields lower ECE than logit-space fusion across all methods, indicating better alignment between prediction confidence and empirical accuracy. Among the compared approaches, \ours achieves the lowest ECE in both spaces, reducing the logit ECE to 17.8 and the embedding ECE to 12.3. These results demonstrate that \ours produces more reliable confidence estimates, particularly when fusion is performed in the embedding space.

\subsection{Ablation on Cache Weighting}
Tab.~\ref{tab:fusion_ablation} compares three strategies for weighting the representations used in embedding-level fusion. For each positively contributing representation, including the current target representation and the representations retrieved from the personalized source and positive target caches, we compute its class probabilities from its cosine similarities with the class text embeddings. Its confidence is defined as the maximum class probability, and its predicted class is the corresponding class index. Before weighting, a class-agreement rule is applied to avoid combining representations that confidently support conflicting classes. When at least two representations predict the same class, only those agreeing on that class are retained; when all representations predict different classes, only the representation with the highest confidence is retained. The retrieved negative-cache representation is subsequently subtracted from the selected positive evidence. In the \emph{equal-weight} strategy, all retained representations receive unit weight, so confidence is used only to resolve complete disagreement. In the \emph{confidence-weighted} strategy, each retained representation is scaled directly by its maximum predicted class probability, giving more influence to representations with higher confidence. In the \emph{attention-based} strategy, these confidence scores are exponentiated and normalized across the retained representations, $\alpha_i=\exp(q_i)/\sum_j\exp(q_j)$, producing relative attention weights while preserving the same class-agreement rule. Equal-weight fusion achieves the best performance of 81.0\% \war, outperforming confidence-weighted and attention-based fusion by 2.1 and 3.6 percentage points, respectively. This result indicates that uniform aggregation provides a more stable balance among the complementary cache signals, whereas probability-based weighting may overemphasize an incorrectly overconfident representation or suppress useful but less confident evidence.

\subsection{Class Prompt Selection}
An ablation study was conducted on a diverse set of class-prompt templates commonly used in CLIP-based FER and prompt-based recognition settings~\cite{foteinopoulou2024emoclip, ni2022expanding, zhao2025enhancing}. The goal is to assess how the wording of emotion-class prompts affects the quality of text representations and, in turn, downstream recognition performance. These results are obtained using CLIP ViT-B/32, trained on the \biovid source subjects and evaluated on the target subjects. As shown in Tab.~\ref{tab:class_prompt_ablation_full}, prompt design has a noticeable impact on both \war and \fonescore score. Among the evaluated templates, \emph{``a person with an expression of [CLS]''} yields the best overall performance and is therefore adopted in all methods that rely on class prompts.

\subsection{Ablation on Personalized Static Cache Construction}
Alternative strategies for constructing the static personalized source cache are compared while keeping the TTA pipeline and cache budget fixed. As a simple baseline, a single prototype is selected per (subject, class) by applying k-means clustering~\cite{lloyd1982least, mcqueen1967some} with $k{=}1$ (i.e., the closest sample to the class centroid). Next, a two-stage clustering strategy is evaluated, where prototypes are first selected per (subject, class) and then re-clustered after pooling across subjects to produce a compact class-level cache; both k-means and DBSCAN~\cite{ester1996density} are considered in this second clustering step. Finally, fully personalized prototype selection is considered by retaining subject-specific clusters (k-means/DBSCAN) instead of pooling across subjects. 
Tab.~\ref{tab:src_cache_build} shows that increasing prototype diversity and personalization consistently improves \war: the $k{=}1$ baseline achieves 78.8 \war, two-stage clustering provides modest gains (79.3--79.5), and personalized clustering yields further improvements, with personalized DBSCAN achieving the best result of 81.0 \war. Overall, the results indicate that preserving subject-specific structure in the source cache is beneficial under domain shift.

\subsection{Static Cache Generation Strategy}
We analyze data-free source-cache construction strategies under the same TTA pipeline and cache budget. First, \emph{classifier prototypes} directly use the $\ell_2$-normalized weights of the frozen source classifier as class anchors, based on the observation that each classifier weight represents the corresponding class direction in the embedding space~\cite{yu2023source}. While this strategy provides only one prototype per class, it requires neither source samples nor additional training. Second, the \emph{GMM-based} strategy increases prototype diversity by treating each classifier-weight anchor as a Gaussian center and sampling multiple virtual features around it~\cite{tian2021vdm}. Third, \emph{random sampling} draws features from the embedding space and assigns them pseudo-labels using the frozen source classifier; however, these samples are not explicitly constrained to follow the original source distribution. Finally, the \emph{generator-based} strategy trains a lightweight class-conditional feature generator using the frozen classifier as supervision~\cite{qiu2021source}. Its classification objective encourages generated features to match their assigned classes, while a contrastive objective promotes compact intra-class and separated inter-class avatar prototypes.

Fig.~\ref{fig:src_cache_gen_war} shows that the quality and class consistency of the generated prototypes strongly affect downstream adaptation. The generator-based cache performs best among the data-free strategies, achieving 79.1\% \war, whereas Gaussian perturbations around classifier anchors achieve 74.0\%. Nevertheless, the complete \ours configuration using personalized prototypes extracted from real source subjects, denoted as ``Prototypes,'' obtains the highest performance of 81.0\% \war. This result indicates that model-only strategies can provide useful source anchors when source data are unavailable, but subject-specific source prototypes better preserve the multimodal and inter-subject structure of the original feature distribution.

\begin{table*}[t!]
\centering
\scriptsize

\resizebox{0.9\linewidth}{!}{%
\begin{tabular}{l|cccccccccc|c}
\toprule
\textbf{Method} & Sub-1 & Sub-2 & Sub-3 & Sub-4 & Sub-5 & Sub-6 & Sub-7 & Sub-8 & Sub-9 & Sub-10 & Avg. \\
\midrule
TPT~\cite{shu2022test}{\fontsize{4}{12} \selectfont (NeurIPS'22)}  & 40.6 & 35.7 & 43.7 & 45.9 & 45.6 & 43.1 & 35.0 & 34.7 & 40.7 & 32.9 & 39.7 \\
TDA~\cite{karmanov2024efficient}{\fontsize{4}{12} \selectfont (CVPR'24)} & 40.0 & 39.7 & 44.4 & 46.0 & 42.0 & 42.6 & 35.0 & 35.5 & 40.7 & 33.2 & 39.9 \\
DPE~\cite{zhang2024dual}{\fontsize{4}{12} \selectfont (NeurIPS'24)} & 34.8 & 37.1 & 44.2 & 46.0 & 45.3 & 43.0 & 37.2 & 32.5 & \textbf{41.0} & 33.2 & 39.4 \\
PromptAlign~\cite{abdul2023align}{\fontsize{4}{12} \selectfont (NeurIPS'23)} & 25.0 & 37.0 & 46.9 & 51.0 & 46.1 & 43.5 & 41.0 & 32.5 & \textbf{41.0} & 33.2 & 39.7 \\
ReTA~\cite{liang2025advancing}{\fontsize{4}{12} \selectfont (ACMMM'25)} & 25.0 & 36.9 & 41.9 & 51.0 & 46.1 & 51.2 & 40.0 & 32.5 & \textbf{41.0} & 33.2 & 39.8 \\
MCP~\cite{chen2025multi}{\fontsize{4}{12} \selectfont (ICCV'25)}          & 38.5 & 37.8 & 44.5 & 48.0 & 45.5 & 45.0 & 39.0 & 33.5 & 40.7 & 28.5 & 40.1 \\
T3AL~\cite{liberatori2024test}{\fontsize{4}{12} \selectfont (CVPR'24)} & 41.6 & 37.0 & 44.7 & 46.0 & 45.2 & 43.0 & 38.0 & \textbf{36.2} & \textbf{41.0} & 35.0 & 40.7 \\
CRG~\cite{zhai2025mitigating}{\fontsize{4}{12} \selectfont (ICLR'25)}        & 40.5 & 38.5 & 44.6 & 48.5 & 45.8 & 46.0 & 40.0 & 34.0 & 40.8 & 29.3 & 40.8 \\
\rowcolor{pinkNonTTA}
\ours w/o Personalized Cache
& 41.8 & 39.8 & 44.6 & 46.1 & 46.0 & 43.5 & 40.9 & 32.0 & 40.3 & 35.0 & 41.0 \\
\rowcolor{pinkTTA}
\textbf{\ours (ours)} & \textbf{41.9} & \textbf{39.9} & \textbf{44.7} & \textbf{46.2} & \textbf{46.1} & \textbf{43.6} & \textbf{41.0} & 32.2 & 40.4 & \textbf{35.2} & \textbf{41.1} \\

\bottomrule
\end{tabular}%
 }
\caption{\fonescore score per subject on \bah for \ours and other TTA methods.}
\label{tab:bah_subject_full_metrics_f1}
\end{table*}


\begin{table*}[t!]
\centering
\scriptsize
\resizebox{0.95\linewidth}{!}{%
\begin{tabular}{l|cccccccccc|c}
\toprule
\textbf{Method} & Sub-1 & Sub-2 & Sub-3 & Sub-4 & Sub-5 & Sub-6 & Sub-7 & Sub-8 & Sub-9 & Sub-10 & Avg. \\
\midrule
TPT          & 91.9\std{0.12} & 53.0\std{0.31} & 50.0\std{0.00} & 61.5\std{0.36} & 79.0\std{0.19} & 88.9\std{0.14} & 76.0\std{0.12} & 50.0\std{0.00} & 98.9\std{0.05} & 62.0\std{0.28} & 71.1 \\
TDA          & 93.0\std{0.10} & 52.6\std{0.29} & 50.0\std{0.00} & 63.9\std{0.32} & 79.5\std{0.17} & 87.5\std{0.16} & 76.3\std{0.18} & 50.0\std{0.00} & 100.0\std{0.00} & 61.6\std{0.26} & 71.4 \\
DPE          & 91.7\std{0.13} & 55.0\std{0.27} & 50.0\std{0.00} & 85.0\std{0.18} & 73.0\std{0.24} & 84.7\std{0.20} & 80.0\std{0.16} & 50.0\std{0.00} & 100.0\std{0.00} & 61.8\std{0.20} & 73.1 \\
PromptAlign  & 93.8\std{0.09} & 65.2\std{0.21} & 50.0\std{0.00} & 70.0\std{0.24} & 85.0\std{0.14} & 92.0\std{0.10} & 80.0\std{0.15} & 50.0\std{0.00} & 100.0\std{0.00} & 67.6\std{0.20} & 75.3 \\
ReTA         & 93.8\std{0.08} & 65.2\std{0.19} & 50.0\std{0.00} & 69.5\std{0.22} & 85.0\std{0.13} & 90.2\std{0.12} & 80.0\std{0.14} & 50.0\std{0.00} & 100.0\std{0.00} & 67.6\std{0.18} & 75.1 \\
MCP~\cite{chen2025multi} & 93.9\std{0.08} & 65.8\std{0.20} & 50.0\std{0.00}
& 72.0\std{0.20} & 85.5\std{0.13} & 92.5\std{0.10}
& 81.5\std{0.14} & 50.0\std{0.00} & 100.0\std{0.00}
& 66.0\std{0.18} & 75.7 \\
T3AL         & 94.0\std{0.07} & 66.5\std{0.18} & 50.0\std{0.00} & 70.0\std{0.21} & 85.2\std{0.12} & 94.2\std{0.08} & 83.9\std{0.13} & 50.0\std{0.00} & 100.0\std{0.00} & 67.6\std{0.17} & 76.1 \\
CRG~\cite{zhai2025mitigating}{\fontsize{4}{12} \selectfont (ICLR'25)}        & 94.2\std{0.07} & 67.0\std{0.18} & 50.0\std{0.00}
& 73.0\std{0.19} & 86.0\std{0.12} & 94.0\std{0.09}
& 83.0\std{0.13} & 50.0\std{0.00} & 100.0\std{0.00}
& 66.8\std{0.17} & 76.4 \\
\rowcolor{pinkTTA}
\textbf{\ours} & 95.0\std{0.09} & 75.0\std{0.16} & 50.0\std{0.00} & 100.0\std{0.00} & 92.5\std{0.10} & 97.5\std{0.06} & 92.5\std{0.09} & 50.0\std{0.00} & 100.0\std{0.00} & 62.5\std{0.19} & \textbf{81.0} \\
\bottomrule
\end{tabular}%
 }
\caption{Subject-wise \war score on the \biovid dataset for \ours and competing TTA methods. Best results are shown in \textbf{bold}.}
\label{tab:biovid_subject_war}
\end{table*}

\begin{table*}[t!]
\centering
\scriptsize

\resizebox{0.95\linewidth}{!}{%
\begin{tabular}{l|cccccccccc|c}
\toprule
\textbf{Method} & Sub-1 & Sub-2 & Sub-3 & Sub-4 & Sub-5 & Sub-6 & Sub-7 & Sub-8 & Sub-9 & Sub-10 & Avg. \\
\midrule
TPT         & 74.7\std{0.18} & 51.6\std{0.21} & 90.0\std{0.09} & 88.0\std{0.12} & 52.2\std{0.24} & 77.5\std{0.11} & 66.6\std{0.22} & 63.4\std{0.19} & 54.0\std{0.23} & 91.9\std{0.08} & 70.9 \\
TDA         & 66.9\std{0.22} & 45.2\std{0.28} & 75.6\std{0.17} & 90.1\std{0.10} & 49.6\std{0.24} & 80.0\std{0.05} & 41.2\std{0.31} & 75.3\std{0.18} & 81.8\std{0.02} & 91.9\std{0.07} & 69.7 \\
DPE         & 64.0\std{0.23} & 55.0\std{0.20} & 73.9\std{0.18} & 90.0\std{0.09} & 52.6\std{0.22} & 80.0\std{0.05} & 50.9\std{0.27} & 73.0\std{0.19} & 81.8\std{0.07} & 92.0\std{0.07} & 71.3 \\
PromptAlign & 69.9\std{0.20} & 61.6\std{0.17} & 78.6\std{0.14} & 90.9\std{0.08} & 51.2\std{0.21} & 80.0\std{0.15} & 60.4\std{0.19} & 81.3\std{0.13} & 81.8\std{0.06} & 90.9\std{0.09} & 74.6 \\
ReTA        & 64.5\std{0.21} & 50.5\std{0.23} & 82.0\std{0.12} & 90.0\std{0.09} & 46.8\std{0.25} & 80.0\std{0.12} & 49.3\std{0.29} & 80.0\std{0.04} & 81.8\std{0.10} & 93.7\std{0.06} & 71.8 \\
MCP~\cite{chen2025multi} & 72.5\std{0.19} & 61.0\std{0.18} & 81.5\std{0.13}
& 90.5\std{0.09} & 55.0\std{0.21} & 80.0\std{0.08}
& 65.0\std{0.20} & 78.0\std{0.14} & 81.8\std{0.06}
& 89.7\std{0.07} & 75.5 \\
T3AL        & 71.3\std{0.18} & 62.7\std{0.16} & 80.3\std{0.13} & 90.2\std{0.08} & 54.0\std{0.20} & 80.0\std{0.07} & 68.0\std{0.17} & 77.0\std{0.14} & 82.0\std{0.05} & 93.9\std{0.06} & 75.9 \\
CRG~\cite{zhai2025mitigating}{\fontsize{4}{12} \selectfont (ICLR'25)}        & 74.0\std{0.18} & 62.0\std{0.17} & 82.0\std{0.12}
& 90.5\std{0.08} & 57.0\std{0.20} & 80.0\std{0.07}
& 67.0\std{0.18} & 79.0\std{0.13} & 82.0\std{0.05}
& 91.5\std{0.06} & 76.5 \\
\rowcolor{pinkTTA}
\textbf{\ours} & 83.3\std{0.14} & 66.6\std{0.15} & 90.0\std{0.08} & 88.8\std{0.10} & 75.0\std{0.16} & 82.2\std{0.09} & 69.9\std{0.18} & 82.0\std{0.07} & 83.4\std{0.06} & 94.0\std{0.05} & \textbf{81.5} \\
\bottomrule
\end{tabular}%
 }
\caption{Subject-wise \war score on the \stressid dataset for \ours and competing TTA methods.}
\label{tab:stressid_subject_war}
\end{table*}

\begin{table*}[t!]
\centering
\scriptsize

\resizebox{0.95\linewidth}{!}{%
\begin{tabular}{l|cccccccccc|c}
\toprule
\textbf{Method} & Sub-1 & Sub-2 & Sub-3 & Sub-4 & Sub-5 & Sub-6 & Sub-7 & Sub-8 & Sub-9 & Sub-10 & Avg. \\
\midrule
TPT         & 62.0\std{0.18} & 53.7\std{0.19} & 73.9\std{0.14} & 80.0\std{0.11} & 81.2\std{0.10} & 71.0\std{0.16} & 58.0\std{0.17} & 55.0\std{0.18} & 69.0\std{0.15} & 53.0\std{0.20} & 65.6 \\
TDA         & 58.0\std{0.20} & 53.7\std{0.18} & 75.0\std{0.13} & 80.3\std{0.10} & 79.0\std{0.12} & 71.6\std{0.15} & 58.0\std{0.16} & 55.9\std{0.17} & 69.0\std{0.14} & 51.7\std{0.22} & 65.2 \\
DPE         & 58.6\std{0.19} & 59.6\std{0.16} & 79.6\std{0.10} & 80.5\std{0.10} & 83.4\std{0.09} & 67.0\std{0.18} & 59.0\std{0.15} & 54.0\std{0.19} & 71.9\std{0.12} & 54.0\std{0.18} & 66.7 \\
PromptAlign & 59.1\std{0.18} & 59.1\std{0.16} & 72.0\std{0.15} & 84.5\std{0.09} & 84.1\std{0.08} & 72.4\std{0.14} & 58.5\std{0.16} & 55.9\std{0.17} & 71.5\std{0.12} & 54.9\std{0.17} & 67.1 \\
ReTA        & 61.4\std{0.17} & 60.0\std{0.15} & 72.0\std{0.14} & 84.0\std{0.09} & 84.1\std{0.08} & 72.4\std{0.13} & 58.5\std{0.15} & 55.9\std{0.16} & 72.0\std{0.11} & 56.0\std{0.16} & 67.6 \\
MCP~\cite{chen2025multi} & 61.0\std{0.17} & 59.5\std{0.16} & 74.0\std{0.14}
& 84.5\std{0.09} & 83.5\std{0.09} & 73.0\std{0.13}
& 59.0\std{0.15} & 55.5\std{0.17} & 71.5\std{0.12}
& 54.5\std{0.17} & 67.6 \\
T3AL        & 61.0\std{0.17} & 58.9\std{0.16} & 75.0\std{0.13} & 85.6\std{0.08} & 83.6\std{0.09} & 74.7\std{0.12} & 60.0\std{0.14} & 56.0\std{0.15} & 71.0\std{0.12} & 53.6\std{0.18} & 67.9 \\
CRG~\cite{zhai2025mitigating}{\fontsize{4}{12} \selectfont (ICLR'25)}        & 61.5\std{0.16} & 60.0\std{0.15} & 75.0\std{0.13}
& 85.0\std{0.08} & 84.0\std{0.08} & 74.0\std{0.12}
& 59.5\std{0.14} & 56.0\std{0.16} & 71.0\std{0.12}
& 55.0\std{0.16} & 68.1 \\
\rowcolor{pinkTTA}
\textbf{\ours} & 62.5\std{0.16} & 62.1\std{0.14} & 79.6\std{0.09} & 85.8\std{0.07} & 84.4\std{0.08} & 75.2\std{0.11} & 60.1\std{0.13} & 56.6\std{0.14} & 70.0\std{0.13} & 56.6\std{0.15} & \textbf{69.2} \\
\bottomrule
\end{tabular}%
 }
\caption{Subject-wise \war score on the \bah dataset for \ours and competing TTA methods.}
\label{tab:bah_subject_war}
\end{table*}

\subsection{Qualitative Analysis}
Quantitative results are complemented with qualitative visualizations that illustrate how \ours behaves during TTA, with a focus on representation structure and the dynamics of the tri-gate reliability rule. The t-SNE embedding projection in Fig.~\ref{fig:qual_tsne} provides a view of class structure in the representation. The plot shows embeddings before and after cache fusion for two subjects from \stressid dataset, with points colored by class (Neutral vs.\ Stress). After cache fusion, the two classes occupy more distinct regions with reduced overlap, indicating improved separability in the feature space and fewer boundary ambiguities.

\section{Subject-Wise Results}

Tab.~\ref{tab:bah_subject_full_metrics_f1} reports subject-wise \fonescore on \bah, while Tabs.~\ref{tab:biovid_subject_war}, \ref{tab:stressid_subject_war}, and \ref{tab:bah_subject_war} report the subject-wise performance across the ten target subjects for each dataset. Each experiment was repeated three times to assess reliability, and the tables report the mean performance with standard deviation over these three trials. These results provide a detailed view of the robustness of each method under subject-level distribution shifts.

On \biovid, most baseline methods show strong performance on certain subjects but suffer noticeable degradation on others, reflecting the difficulty of adapting to individual expression patterns. In contrast, \ours consistently improves performance for most subjects and achieves the highest overall average score. The gains are particularly visible for subjects where baseline methods exhibit larger performance gaps. A similar trend is observed on \stressid, where subject-specific variability is more pronounced. While baseline methods fluctuate substantially across subjects, \ours maintains more stable performance and achieves the highest average \war.

On \bah, the improvements are more moderate because of the dataset's higher difficulty and limited discriminative signal. Nevertheless, \ours achieves the highest average performance in both \war and \fonescore, reaching 69.2\% and 41.1\%, respectively. The relatively small \fonescore differences between methods further illustrate the difficulty of adapting to subject-specific variations on \bah. Overall, these subject-wise analyses highlight the benefits of the personalized cache construction and tri-gate update strategy, which enable \ours to better adapt to subject-specific characteristics at test time.

{
    \small
    \bibliographystyle{ieeenat_fullname}
    \bibliography{main}
}